\documentclass[10pt,twocolumn,letterpaper]{article}

\usepackage{cvpr}              

\usepackage{comment}
\usepackage{amsmath,amssymb} 
\usepackage{color}

\usepackage{graphics}
\usepackage{algpseudocode}
\usepackage{fontawesome}
\usepackage{colortbl}
\usepackage{dsfont}
\usepackage{stackengine}
\usepackage{times}
\usepackage{epsfig}
\usepackage{graphicx}
\usepackage{amsmath}
\usepackage{amssymb}
\usepackage{multirow}
\usepackage{mathtools,xparse}
\usepackage{subcaption}
\usepackage{stmaryrd}
\usepackage{placeins}
\usepackage{pifont}
\usepackage{tabularx}
\usepackage{mdframed}
\usepackage{makecell}
\usepackage{arydshln}
\usepackage{booktabs}
\usepackage{enumitem}
\usepackage[skip=1ex,font=small,labelsep=period]{caption}
\usepackage{tabu}
\usepackage{etoolbox}
\usepackage{verbatim}
\usepackage{color}
\usepackage{siunitx}
\usepackage{xspace}
\usepackage{xcolor}

\usepackage{wrapfig}
\usepackage{array}

\usepackage[pagebackref,breaklinks,colorlinks]{hyperref}

\usepackage[capitalize]{cleveref}
\crefname{section}{Sec.}{Secs.}
\Crefname{section}{Section}{Sections}
\Crefname{table}{Table}{Tables}
\crefname{table}{Tab.}{Tabs.}

\newcommand{\mat}[1]{\mathbf{#1}}     

\newcommand{\trans}{\text{T}}

\renewcommand{\d}[1]{\mbox{\boldmath$#1$}}

\newcommand{\vunn}{\d{\theta}}
\newcommand{\unn}{\theta}

\newcommand{\arr}[2]{\begin{array}{#1} #2\end{array}}
\newcommand{\mx}[2]{\left[\!\!\arr{#1}{#2}\!\!\right]}
\newcommand{\s}[1]{\underline{#1}}
\newcommand{\smat}[1]{\s{\mat {#1}}}
\newcommand{\transs}[0]{{'}^{\sf T}}        
\newcommand{\transi}[0]{^{-\sf T}}          

\newcommand{\qr}{\mbox{qr}}
\newcommand{\svd}{\mbox{svd}}

\def\cvprPaperID{6489} 
\def\confName{CVPR}
\def\confYear{2023}

\begin{document}

\title{A Large-Scale Homography Benchmark}

\author{Daniel Barath$^1$, Dmytro Mishkin$^2$, Michal Polic$^{2,3}$, Wolfgang Förstner$^4$, Jiri Matas$^2$\\
$^1$Computer Vision and Geometry Group, ETH Zurich, Switzerland, \\
$^2$Visual Recognition Group, FEE, CTU in Prague, Czech Republic, \\
$^3$CIIRC, CTU in Prague, Czech Republic, \\
$^4$University Bonn, Germany\\
}
\maketitle

\begin{abstract}
    We present a large-scale dataset of Planes in 3D, Pi3D, of roughly 1000 planes observed in 10 000 images from the 1DSfM dataset,  
    and HEB, a large-scale homography estimation benchmark leveraging Pi3D. 
    The applications of the Pi3D dataset
     are diverse, e.g.\ training or evaluating monocular depth, surface normal estimation and image matching algorithms. 
    The HEB dataset consists of 226 260 homographies and includes roughly 4M correspondences.
    The homographies link images that often undergo significant viewpoint and illumination changes. 
    As applications of HEB, 
    we perform a rigorous evaluation of a wide range of robust estimators and deep learning-based correspondence filtering methods, establishing the current state-of-the-art in robust homography estimation. 
    We also evaluate the uncertainty of the SIFT orientations and scales w.r.t.\ the ground truth coming from the underlying homographies and provide codes for comparing uncertainty of custom detectors.
    The dataset is available at \url{https://github.com/danini/homography-benchmark}.
\end{abstract}

\section{Introduction}
\label{sec:intro}

The planar homography is a projective mapping between images of co-planar 3D points. 
The homography induced by a plane is unique up to a scale and has eight degrees-of-freedom (DoF). 
It encodes the intrinsic and extrinsic camera parameters and the parameters of the underlying 3D plane.

The homography plays an important role in the geometry of multiple views~\cite{hartley2003multiple} with hundreds of papers published in the last few decades about its theory and applications.
\begin{figure}[t]
  \centering
    \includegraphics[width=0.47\columnwidth, trim=23mm 17mm 23mm 8mm, clip]{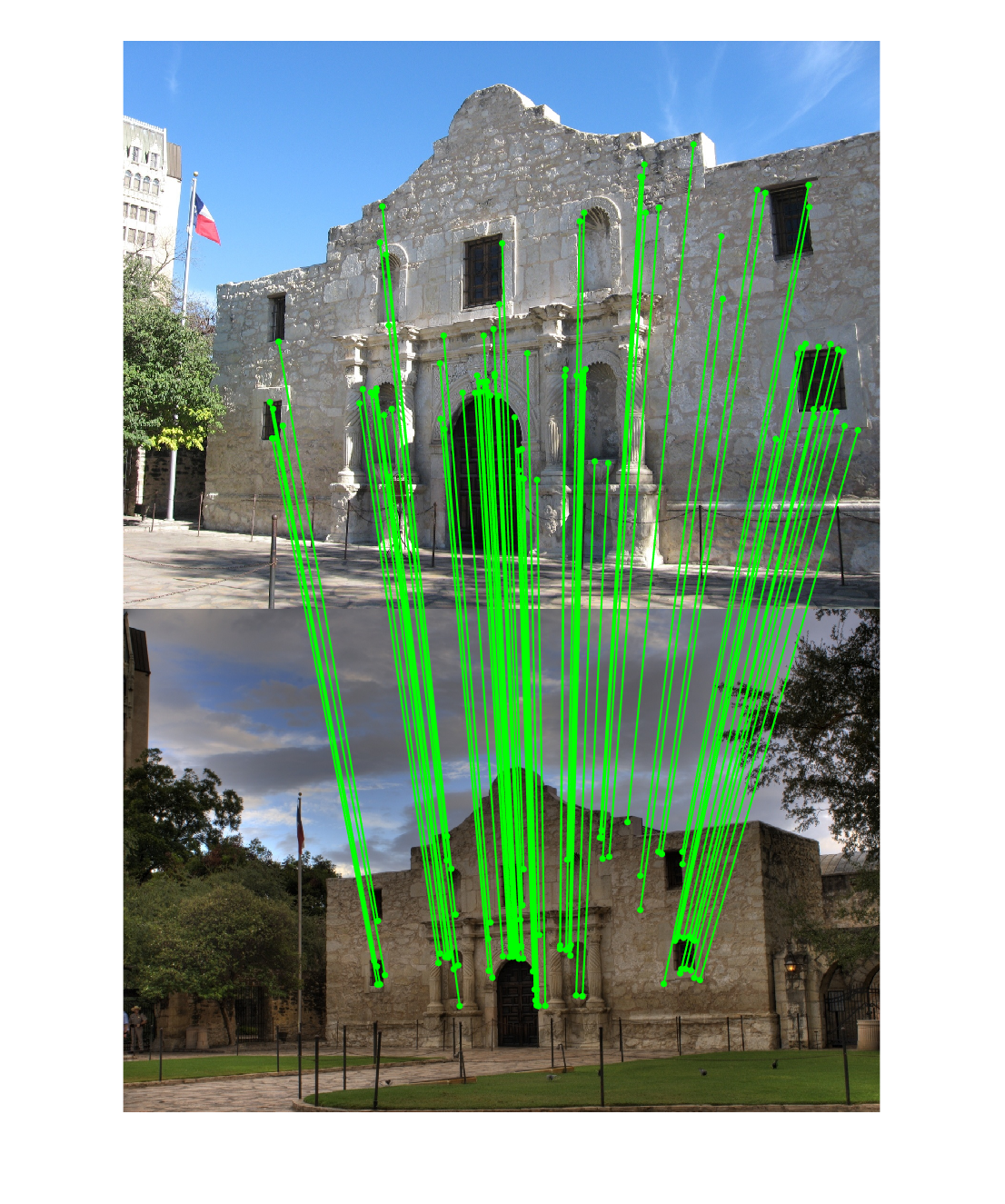}
    \includegraphics[width=0.47\columnwidth, trim=23mm 23.5mm 23mm 13mm, clip]{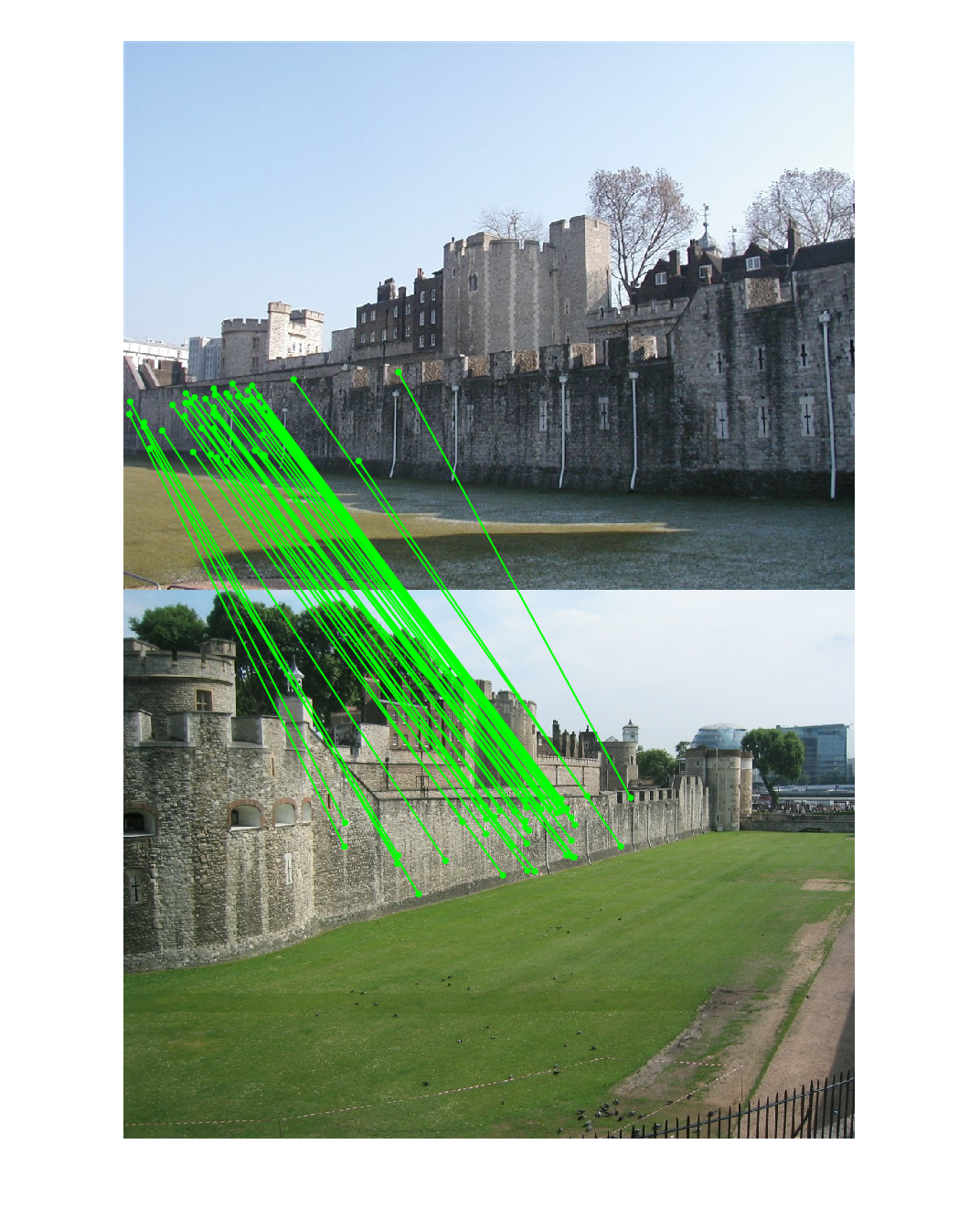}
  \caption{Example image pairs and homographies with their inlier correspondences shown, from the proposed Homography Estimation Benchmark (HEB) dataset. Outliers are not drawn.}
  \label{fig:example_images}
  \vspace{-0.5em}
\end{figure}
%
Estimating planar homographies from image pairs is an important task in computer vision with a number of applications. 
For instance, monocular SLAM systems~\cite{taketomi2017visual,younes2017keyframe,saputra2018visual} rely on homographies when detecting pure rotational camera movements, planar scenes, and scenes with far objects. 
As a homography induced by a plane at infinity represents rotation-only camera motion, it is one of the most important tools for stitching images~\cite{brown2007automatic,adel2014image}.
The generated images cover a larger field-of-view and are useful in various applications, \eg image-based localization~\cite{Arth2011}, SLAM~\cite{Lemaire2007,Ji2020}, autonomous driving~\cite{Wang2020}, sport broadcasting~\cite{chen2018two}, surveillance~\cite{yang2019panoramic}, and augmented and virtual reality~\cite{jethwa1998real,macquarrie2017cinematic}.
Homographies play an important role in  calibration~\cite{zhang2000flexible,chuan2003planar}, metric rectification~\cite{collins1993matching,liebowitz1998metric}, augmented reality~\cite{simon2000markerless,zhou2012robust}, optical flow based on piece-wise planar scene modeling~\cite{yang2015dense}, video stabilization~\cite{grundmann2012calibration,zhou2013plane}, and incremental~\cite{schonberger2016structure} and global~\cite{theia-manual,moulon2016openmvg} Structure-from-Motion.


\begin{figure*}[t]
  \centering
	\begin{subfigure}[t]{0.47\columnwidth}
        \includegraphics[width=1.00\columnwidth, trim=0mm 0mm 0mm 0mm, clip]{assets/graf1.png}\\
        \includegraphics[width=1.00\columnwidth, trim=0mm 0mm 0mm 0mm, clip]{assets/graf2.png}
        \caption{Homogr Dataset}
	\end{subfigure}\hspace{2mm}
	\begin{subfigure}[t]{0.47\columnwidth}
        \includegraphics[width=1.00\columnwidth, trim=14.5mm 0mm 10mm 0mm, clip]{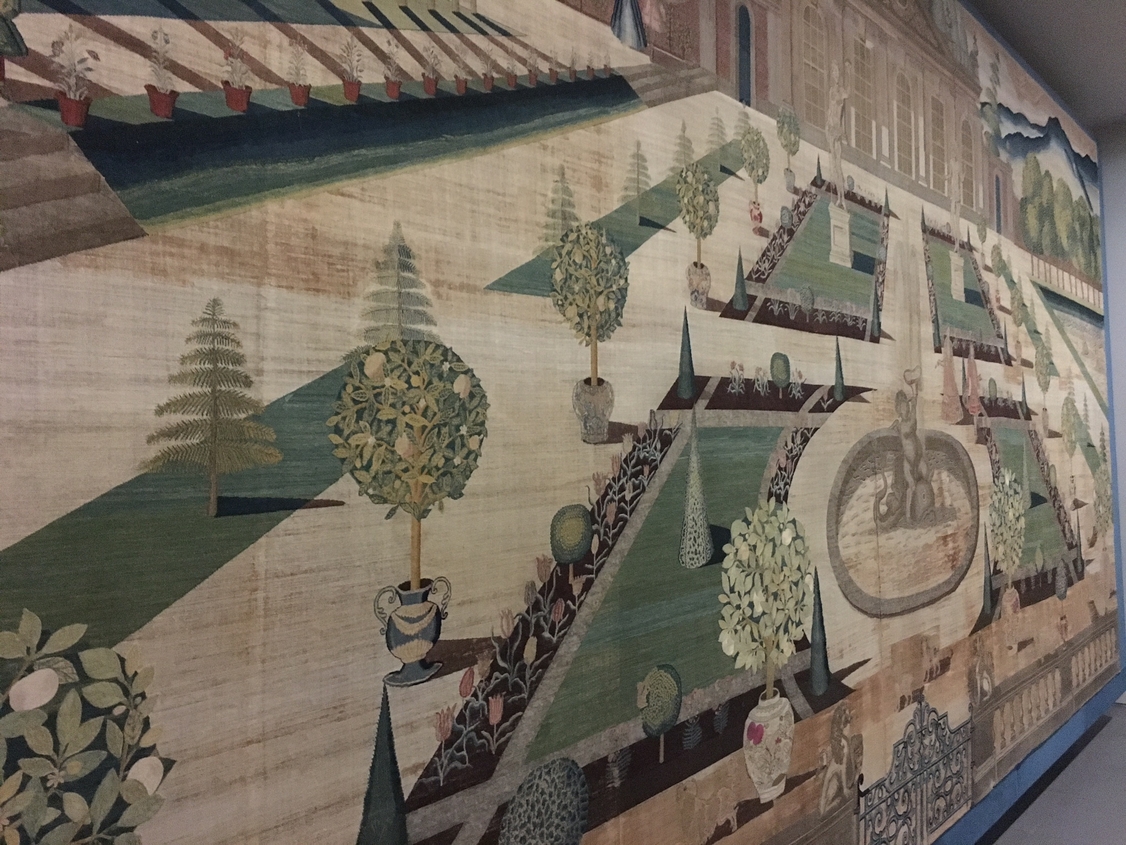}\\
        \includegraphics[width=1.00\columnwidth, trim=14mm 0mm 10mm 0mm, clip]{assets/hpatches_gardens_4.jpg}
        \caption{HPatches Dataset}
	\end{subfigure}\hspace{2mm}
	\begin{subfigure}[t]{0.47\columnwidth}
        \includegraphics[width=1.00\columnwidth, trim=30mm 0mm 28mm 0mm, clip]{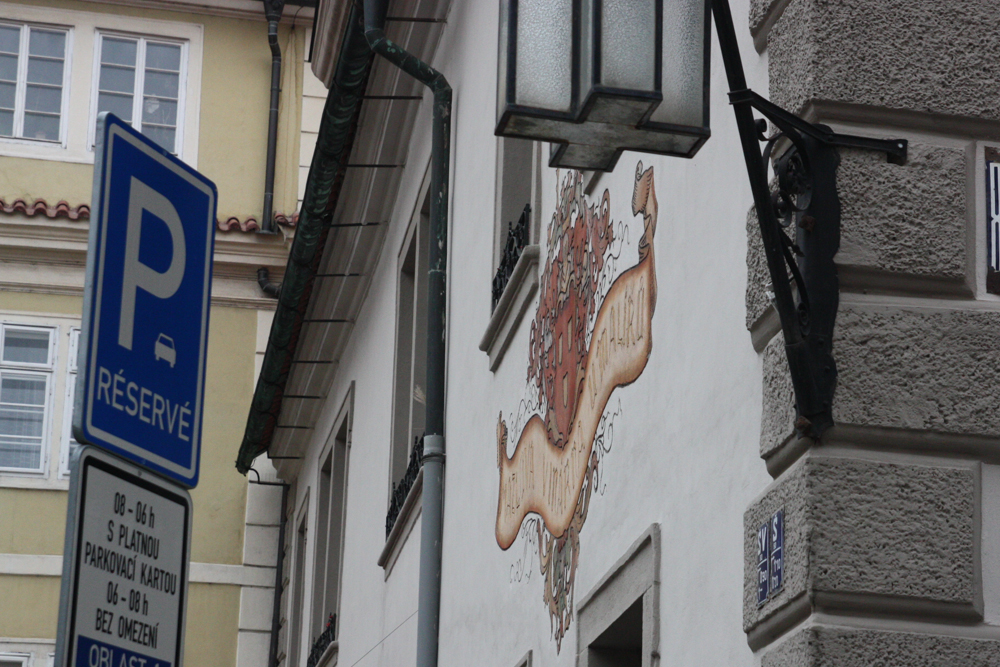}\\
        \includegraphics[width=1.00\columnwidth, trim=20mm 0mm 15mm 0mm, clip]{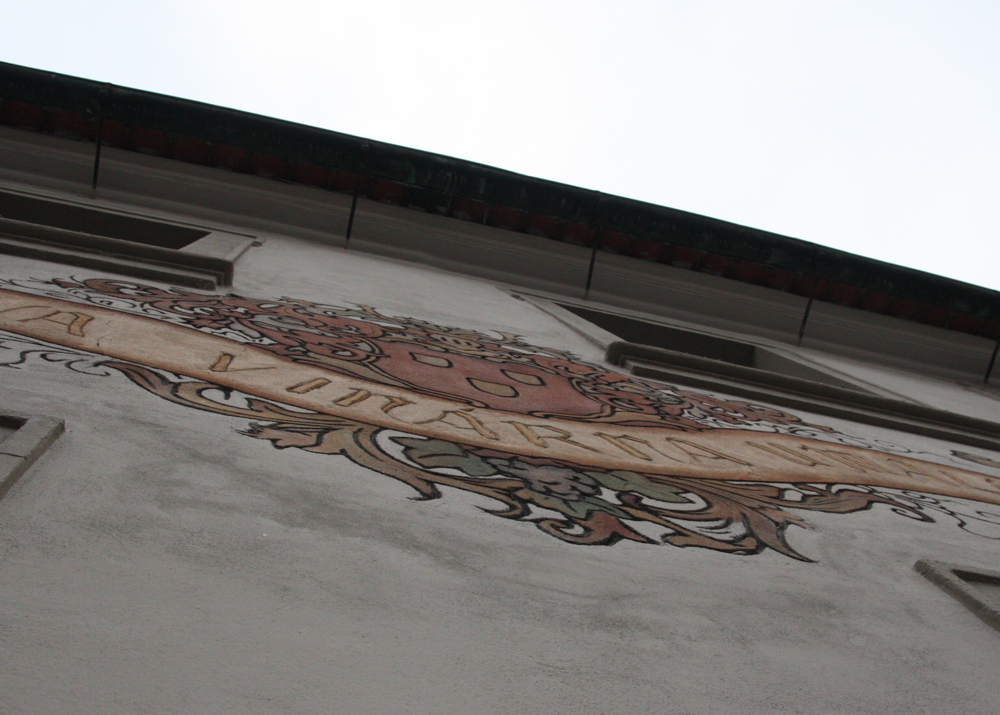}
        \caption{ExtremeView Dataset}
	\end{subfigure}\hspace{2mm}
	\begin{subfigure}[t]{0.47\columnwidth}
        \includegraphics[width=1.00\columnwidth, trim=11mm 0mm 80mm 0mm, clip]{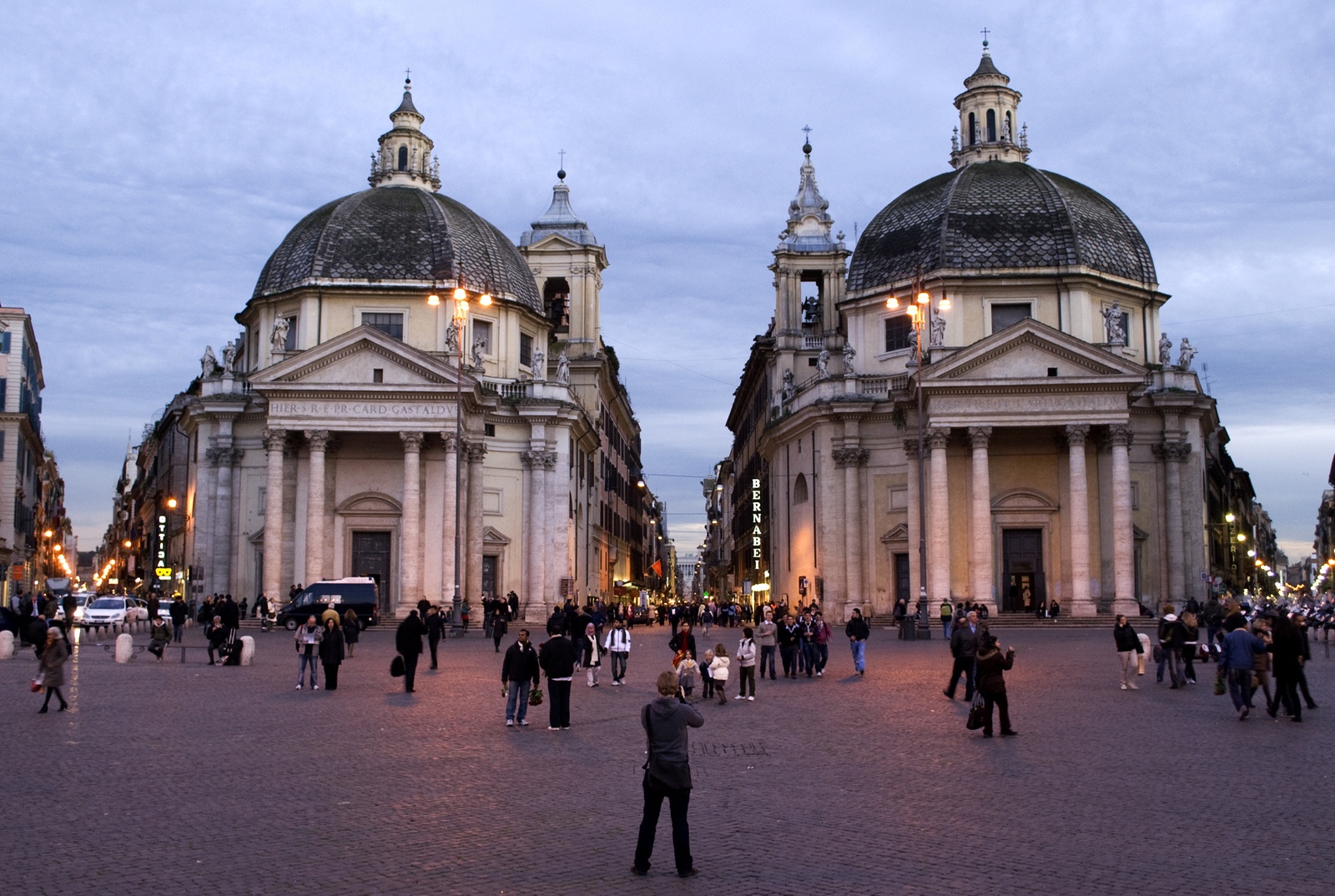}\\
        \includegraphics[width=1.00\columnwidth, trim=11mm 0mm 80mm 0mm, clip]{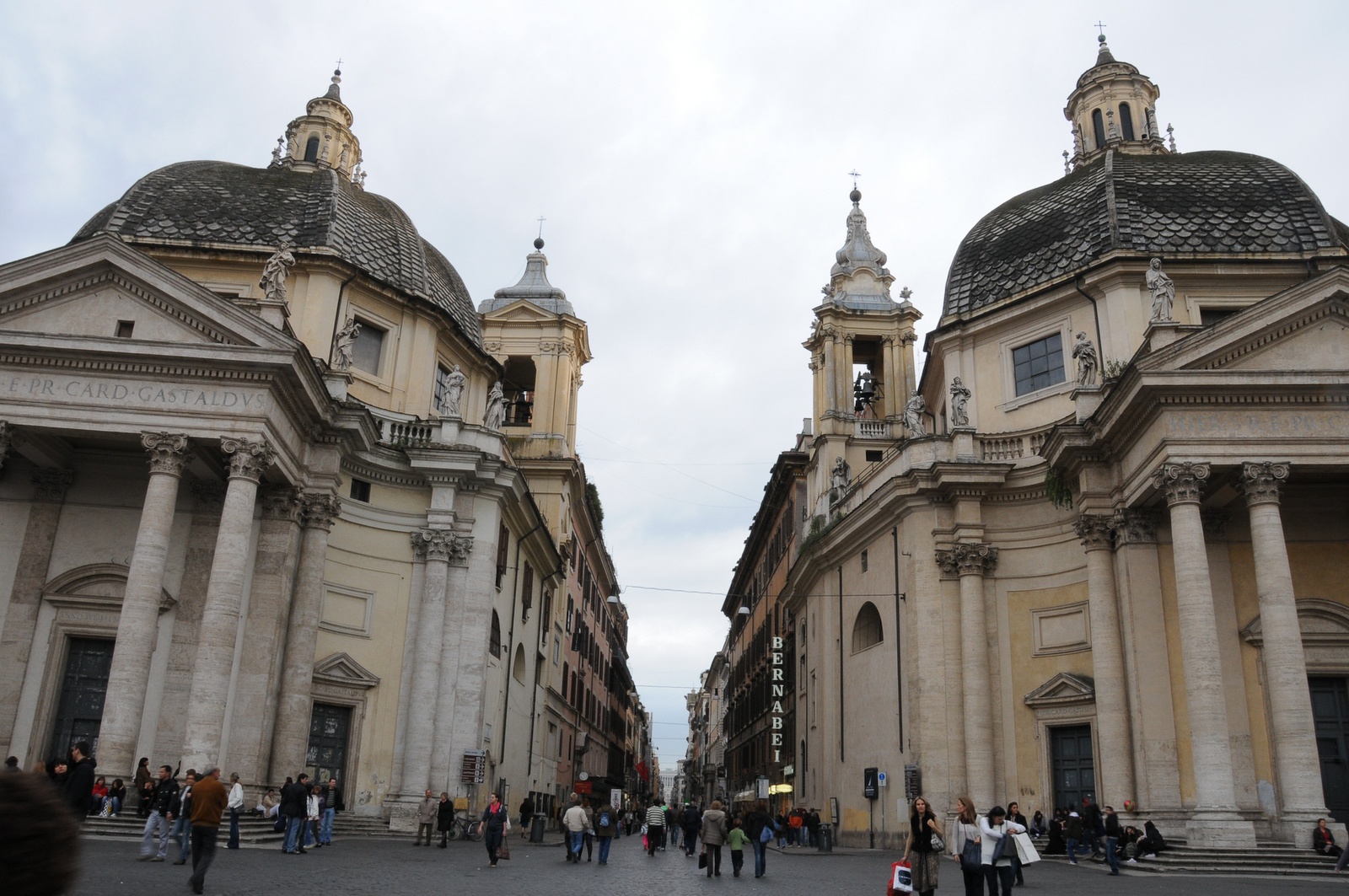}
        \caption{Proposed HEB Dataset}
	\end{subfigure}
  \caption{ Typical image pairs (a-c) from widely used datasets for homography estimator benchmarking and (d) from HEB.
 }
  \label{fig:example_images_from_datasets}
\end{figure*}

The traditional approach of finding homographies in image pairs consists of two main stages.
First, similarly as in most algorithms working with pairs, feature points are detected and matched~\cite{SIFT2004,ORB2011,brown2007automatic,schoenberger2016vote,IMC2020}. 
They are then often filtered by the widely-used second nearest neighbors (SNN) ratio~\cite{lowe1999object,SIFT2004} or by deep learned filtering methods~\cite{cne2018,acne2020,dfe2018,clnet2021}, to remove gross outliers and, therefore, improve the robust estimation procedure that follows. 
The found tentative point correspondences are contaminated by various sources of noise due to, \eg, measurement and quantization, and a large proportion of them are still outliers -- correspondences inconsistent with the sought model manifold. 
Consequently, some form of robust estimation has to be applied to find a set of inliers and to estimate the parameters of the sought homography.
In practice, either a randomized RANSAC-like~\cite{fischler1981random} robust estimator or an iteratively re-weighted least squares fitting~\cite{holland1977robust} is applied.

The number of datasets on which recent homography and, in general, robust estimation papers evaluate their algorithms is severely limited. 
The Homogr dataset~\cite{fixingLORANSAC2012} consists only of a few image pairs with relatively small baselines and, thus, high inlier ratios.
Given that recent robust estimators, \eg \cite{barath2018graph}, report lower than $0.5$ pixel average re-projection errors on the provided manually labeled correspondences, it is safe to say that this dataset is solved.
The HPatches dataset~\cite{hpatches2017} consists of a few hundreds of image pairs, all looking at an almost completely planar scene, with either significant illumination or viewpoint (mostly in tilt angle) changes. 
While \cite{hpatches2017} is a useful tool for evaluating local feature detector and image matching methods, it is very easy for robust estimators~\cite{cvpr2020ransactutorial}.
The ExtremeView (EVD) dataset~\cite{Mishkin2015MODS} poses a significantly more challenging problem for homography estimation than the previous two. 
The images undergo extreme view-point changes, therefore making both the feature matching and robust estimation tasks especially challenging. However, EVD consists only of 15 image pairs, severely limiting its benchmarking power. 

Besides the data part, a good benchmark has well-defined parameter tuning (training) and evaluation protocols and training-test set split. 
Otherwise, as it happens in other fields, the seemingly rapid progress might be an artifact of tuning the algorithms on the test data, or an artifact of the flawed evaluation procedure~\cite{musgrave2020metric,pGT2021,goyal2021revisiting}.

In short, there are no available large-scale benchmarks with ground truth (GT) homographies that allow evaluating new algorithms on standard internet photos, \ie, ones not necessarily looking at completely planar scenes.  

As the \textit{first contribution}, we create a large-scale dataset of \num{1046} large Planes in 3D (Pi3D) from a standard landmark dataset~\cite{wilson_eccv2014_1dsfm}. 
We use the scenes from the 1DSfM dataset as input and find 3D planes in the reconstructions. 
%
\textit{Second}, we use the Pi3D dataset to find image pairs with estimatable homographies and create a large-scale homography benchmark (HEB) containing a total of \num{226260} homographies that can be considered GT when testing new algorithms (see Fig.~\ref{fig:example_images} for examples). 
A large proportion of the image pairs capture significant viewpoint and illumination changes.
The homographies typically have low inlier ratio, thus making the robust estimation task challenging.
\textit{Third}, we compare a wide range of robust estimators, including recent ones based on neural networks, establishing the current state-of-the-art in robust homography estimation. 
As the \textit{forth} contribution, we demonstrate that the dataset can be used to evaluate the uncertainty of partially or fully affine covariant features detectors~\cite{SIFT2004,AffNet2018}.
While we show it on DoG features~\cite{lowe1999object}, the homographies can be leveraged similarly for the comparison with other detectors. 

\begin{table*}[t!]
\centering
    \resizebox{1.0\linewidth}{!}{\begin{tabular}{ l | c c c c c c c }
    \hline   
        \rowcolor{black!10}
        Dataset & \# image pairs & train-test split & camera pose & scene type & baseline & illumination change & inlier ratio\\ 
    \hline   
        Homogr \cite{fixingLORANSAC2012} & 16 &\ding{55} & \ding{55} & buildings & short/medium & \ding{55} & high\\ 
        ExtremeView \cite{Mishkin2015MODS} & 15 &\ding{55}  & \ding{55} & walls & large & \ding{55} & low\\ 
        HPatches \cite{hpatches2017} & $(59 + 57)\times 5$ & \ding{51}& \ding{55} & walls & short/medium  & \ding{55} $+$ \ding{51} & high \\
        \textbf{HEB} & \num{226260} & \ding{51} &\ding{51} & landmark photos & diverse & \ding{51} & low\\ 
    \hline   
\end{tabular}}
\caption{Comparison of the existing and the proposed HEB homography estimation datasets.} 
\label{tab:datasets}
\end{table*}

\noindent
\textbf{Existing Datasets.}
The datasets traditionally used for evaluating homography estimators are the following. 
%
The \textbf{Homogr} dataset~\cite{fixingLORANSAC2012} consists of 16 image pairs with GT homographies. The GT comes from (also provided) hand-labeled correspondences, which later were optimized to improve the localization accuracy. There is no train-test split, nor a benchmark protocol.
The \textbf{ExtremeView} dataset~\cite{Mishkin2015MODS} consists of 15 image pairs, taken under extreme viewpoint change, together with GT homographies and correspondences. The homographies are derived from hand-labeled correspondences that stem from multiple local feature detectors paired with an affine view synthesis procedure~\cite{Mishkin2015MODS} and RootSIFT descriptor~\cite{RootSIFT2012} matching.
There is no train-test split, nor a benchmark protocol.
The \textbf{HPatches} dataset~\cite{hpatches2017} was introduced in form of local patches for benchmarking descriptors and metric learning methods, later extended with images and homographies. 
%
It consists of 57 image sextuplets with significant illumination but negligible viewpoint changes and 59 ones with viewpoint, but no illumination changes. 
The viewpoint difference mostly consist of tilt (perspective change) in the horizontal direction and some shift -- no big rotation or scale changes.
The GT was obtained from manually annotated correspondences for the initial model estimation and polished by minimizing MSE of image pixel re-projections.
There is no official protocol, nor standard correspondences for homography evaluation -- every paper uses slightly different evaluations, but there is an official train-test split.

To conclude, there is no difficult-enough, large-scale dataset with train-test split and benchmark protocol for evaluating robust homography estimation. 
Table~\ref{tab:datasets} summarizes the properties of each publicly available dataset and, also, that of the proposed one. 
Typical image pairs from the datasets are shown in the first three columns of Fig.~\ref{fig:example_images_from_datasets}.

\begin{figure}[t]
  	\centering
	\begin{subfigure}[t]{0.495\columnwidth}
        \includegraphics[width=1.00\columnwidth, trim=2mm 0mm 2mm 0mm, clip]{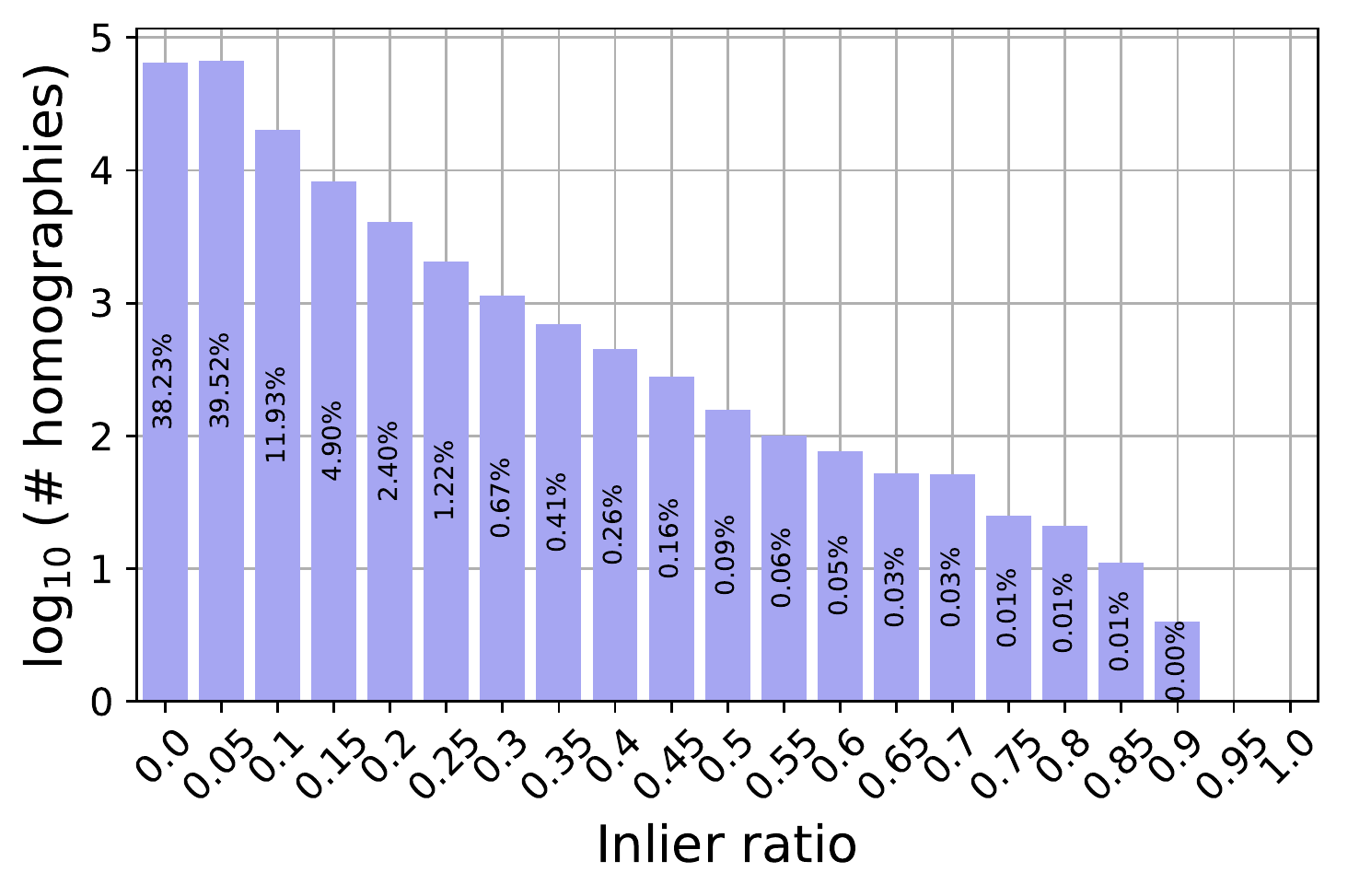}\\[2mm]  
        \includegraphics[width=1.00\columnwidth, trim=2mm 0mm 2mm 0mm, clip]{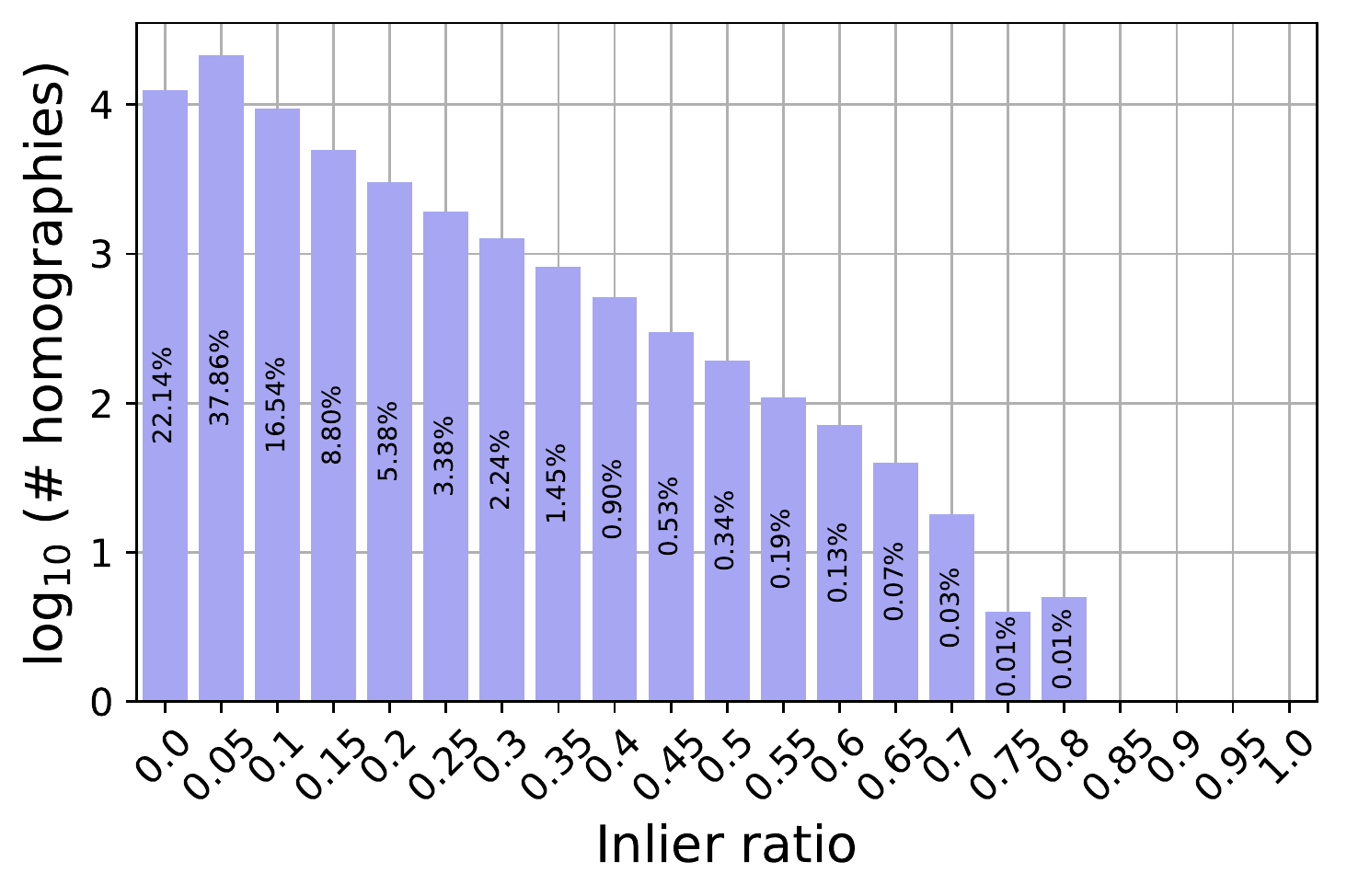}
	\end{subfigure}
	\begin{subfigure}[t]{0.495\columnwidth}
        \includegraphics[width=1.00\columnwidth, trim=2mm 0mm 2mm 0mm, clip]{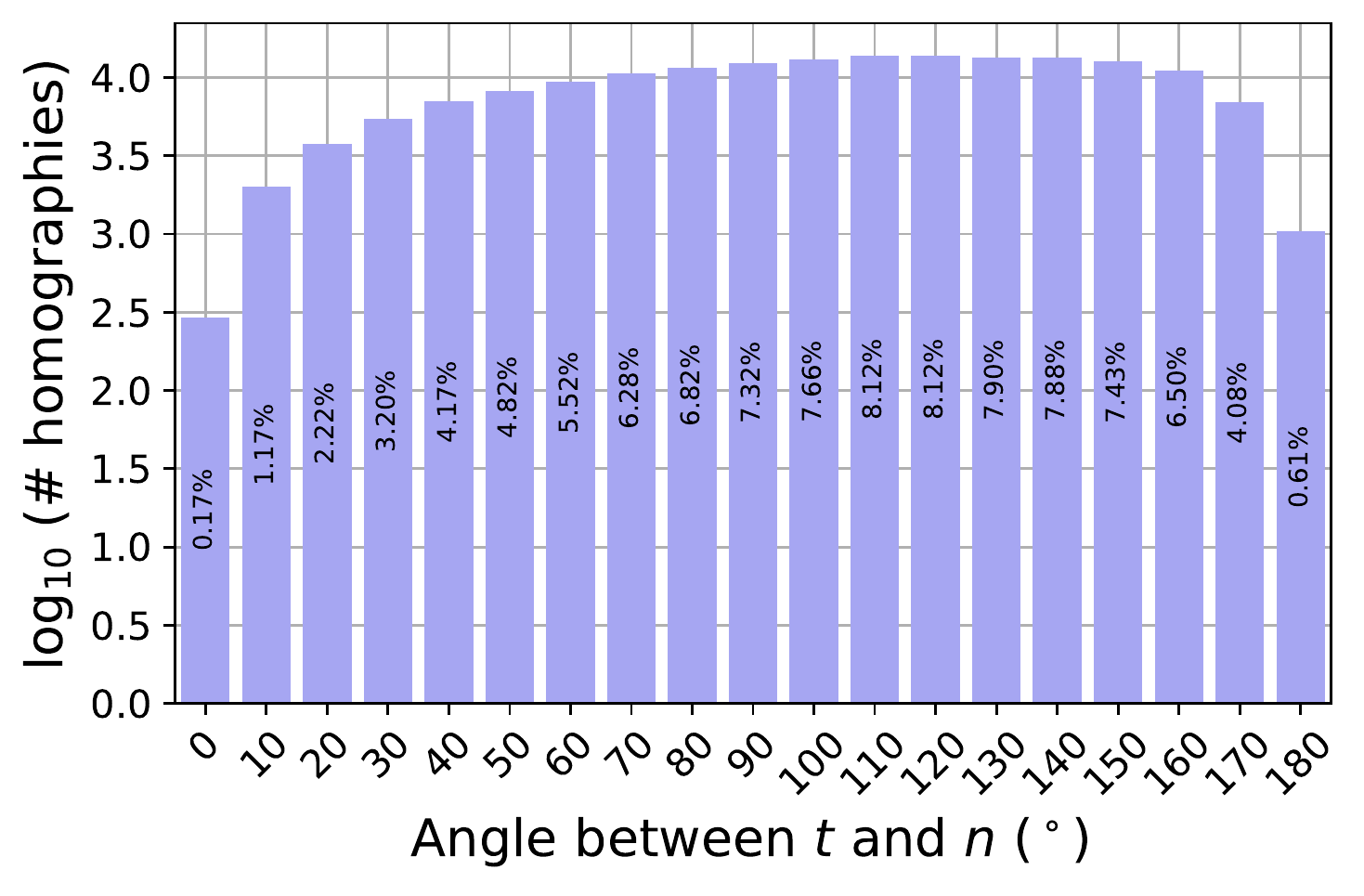}\\[2mm]  
        \includegraphics[width=1.00\columnwidth, trim=2mm 0mm 2mm 0mm, clip]{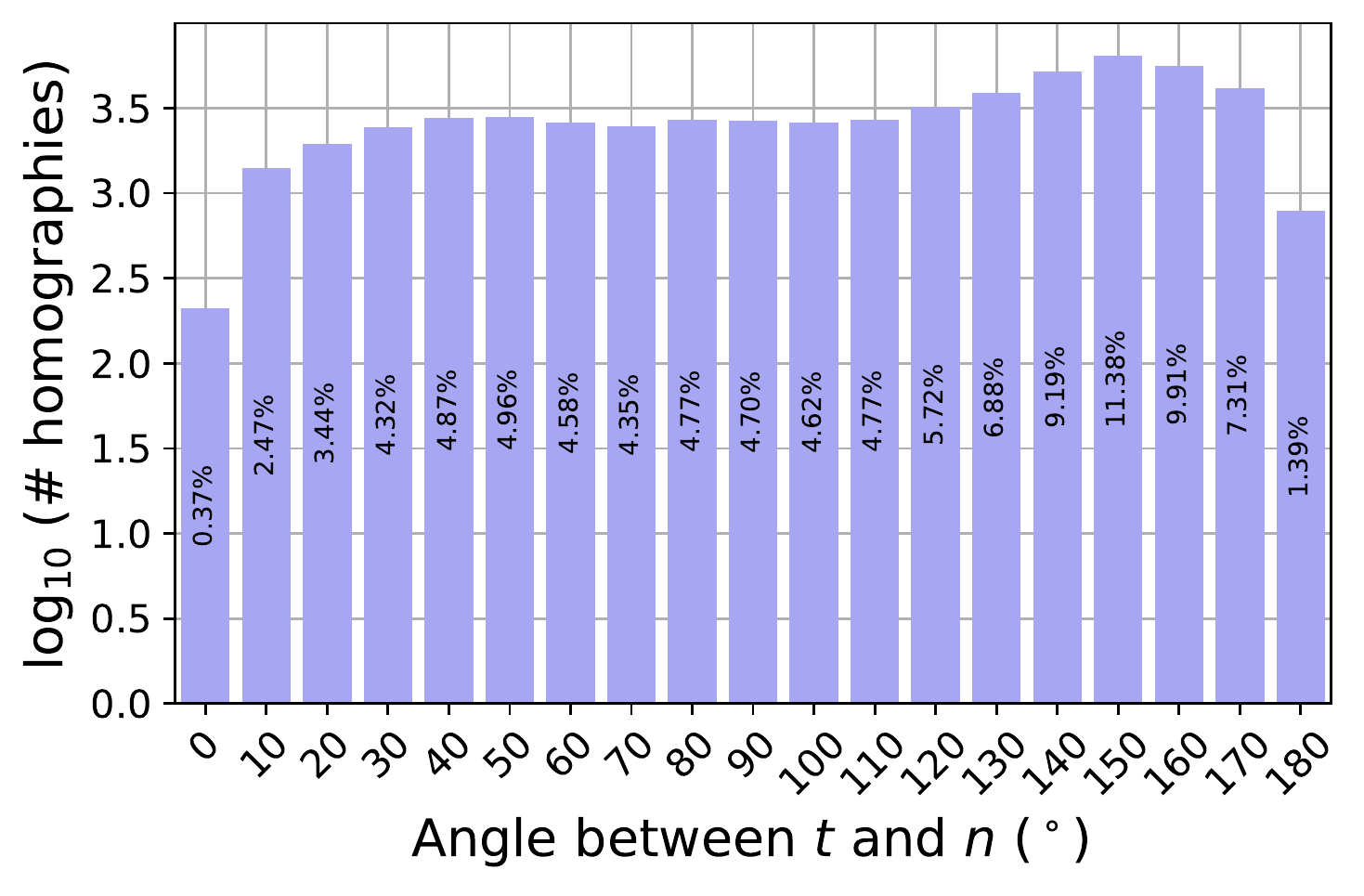}
	\end{subfigure}
	\caption{ \textbf{HEB properties}: test (top; 169 654 pairs) and training (bottom; 56 593 pairs) splits. 
	Percentages are written inside the bars.
	When calculating the angle between the translation and plane normal, 
	the sign of the normal is set so it looks towards the camera. }
    \label{fig:visualization_1}
     \vspace{-0.5em}
\end{figure}

\section{Planes in 3D Dataset}

The Planes in 3D Dataset is based on images from the 1DSfM dataset~\cite{wilson_eccv2014_1dsfm}.
The objective of this section is to create a large-scale dataset of 3D planes in scenes consisting of thousands of real-world photos. 
1DSfM consists of \num{13} scenes of landmarks with photos of varying sizes collected from the internet. 
It provides 2-view matches with epipolar geometries and a reference reconstruction from incremental SfM (with Bundler~\cite{snavely2006photo,snavely2008modeling}) for measuring error. 
Instead, we reconstructed the scenes with COLMAP~\cite{schonberger2016structure}, providing more accurate reconstruction~\cite{knapitsch2017tanks}. 
%
Incremental SfM (\eg, COLMAP) results are often considered GT, \eg in IMC~\cite{IMC2020}, as they are the best which we can get from internet images. 
We manually checked all reconstructions ensuring that only those scenes are used where COLMAP returned an accurate and coherent reconstruction.
We, thus, excluded Gendarmenmarkt and Trafalgar.

We considered several options (\eg, IMC~\cite{IMC2020} and MegaDepth~\cite{li2018megadepth}) before deciding to use 1DSfM. 
We chose it since, nowadays, it is rarely used in computer vision, likely, due to the attached Bundler reconstruction (we replaced it with COLMAP). 
Thus, introducing it back to the community is preferable to keep the variety of commonly used datasets, and not overfitting to IMC, which is only twice bigger than 1DSfM. 
We will make the tools publicly available and, thus, similar data can be easily obtained from other datasets or different features.

Let us introduce the concept of ``estimatable homographies''.
An ``estimatable homography'' is a homography that links two views of a real 3D planar surface; it is consistent with the camera motion; and it is estimatable from its GT correspondences by the standard normalized DLT algorithm~\cite{hartley2003multiple}. 
We keep only those planes in the Pi3D dataset that imply at least a single estimatable homography -- planes that are visible and estimatable in at least an image pair. 
The steps of the pipeline finding such planes and homographies:
\begin{enumerate}[topsep=0.5pt, partopsep=0.5pt,itemsep=0.5pt,parsep=0.5pt]
    \item COLMAP reconstructs the scene from the images.
    \item Multiple 3D planes are detected in the COLMAP point cloud reconstruction.
    \item For each 3D plane, all image pairs  where the plane is visible are selected.
    \item A homography is estimated from each 3D plane in each image pair, where it is visible, using the camera parameters, \ie, the poses and intrinsic matrices.
    \item A homography is rejected if it can not be estimated from only the assigned GT point correspondences, without the camera parameters, accurately. 
\end{enumerate}

\noindent\textbf{Multiple Planes in the Reconstruction.}
The first step of the pipeline is to find 3D planes that can be used when finding planar regions in image pairs. 
For this purpose, we use the Progressive-X$^+$ algorithm~\cite{barath2021progressive}.
To ensure that only dominant planes are found in the reconstruction, we use the following parameters: $n_{\text{min}} = 5000$ and $\epsilon_\text{T} = 0.1$.
Parameter $n_{\text{min}}$ is the number of inliers a plane needs to be considered as a dominant one. 
Parameter $\epsilon_\text{T}$ is the threshold for the pair-wise Tanimoto similarity of the plane consensus vectors. 
Briefly, the Tanimoto similarity measures how similar two planes are in terms of their support. 
These parameters lead to plane segmentations with keeping only the dominant structures and suppressing small details.

\noindent\textbf{Recovering Absolute Scale.}
The COLMAP reconstruction is scaleless, \ie, the metric size of the scene is unknown.
This is why prior work, \eg \cite{IMC2020}, use angle-based metrics to compare camera translations recovered by image matching algorithms. 
Instead, we manually added the scale to the reconstructions. 3D points were re-projected on the images and a manual annotator picked those which are easily identifiable and far enough from each other, \eg, the facade edges of the largest building. We then measured the distance with the ruler tool of Google Maps~\cite{googlemaps}. 
The ratio between these two gives the scaling coefficient to the 3D reconstruction. This procedure is repeated several times and the coefficients are averaged to get the final scale.
The standard deviation of the manually picked absolute scales is approximately $16$ cm, implying that the recovered scales are accurate.

\noindent\textbf{Visible 3D Planes.}
%
%
First, we iterate through all possible image pairs $(I_i, I_j)$, $i, j \in [0, p)$, from the COLMAP reconstruction of the scene, where $p = \binom{n}{2}$ and $n \in \mathbb{N}$ is the number of images. 
For each pair, we collect the planes that have more than ten 3D points visible in both views according to COLMAP depth maps.
%
Second, we detect SIFT features~\cite{SIFT2004} as implemented in OpenCV~
\cite{opencv_library} with RootSIFT~\cite{RootSIFT2012} descriptors. 
In each image, at most $8000$ keypoints are detected and matched. 
We combine mutual nearest neighbor check  to establish tentative point correspondences, as it is recommended in~\cite{IMC2020}. 
The SNN ratio is stored, but no correspondences are filtered out, because different robust estimators, either deep or traditional, may prefer different ratios to achieve their best performance. 

Relative poses are calculated as $\mat R = \mat R_2 \mat R_1^\text{T}$ and $\mat t = \mat t_2 - \mat R_2 \mat R_1^\text{T} \mat t_1$, where $\mat R_1, \mat R_2 \in \text{SO}(3)$ are the absolute rotations and $\mat t_1, \mat t_2 \in \mathbb{R}^3$ are the translations from the reconstruction.
The parameters of the normalized homography implied by the plane are calculated as follows: 
%
    $\mat H = \mat R - (\mat t \mat n^\text{T}) / d$,
%
where $\mat n \in \mathbb{R}^3$ is the plane normal and $d$ is its intercept. 
%
Correspondences are considered inliers if the re-projection error is less than $\epsilon$ pixels given homography $\mat H$.
Homographies with fewer than $10$ inliers are rejected. 
To make sure that the GT homography can be recovered from its inliers and they are not in a degenerate configuration, we estimate homography $\mat H'$ by the normalized DLT algorithm from the inliers.
It is decomposed to rotation $\mat R'$ and translation $\mat t'$ by the standard procedure~\cite{malis2007deeper}. 
We reject homography $\mat H$ if either $\epsilon_{\mat{R}'} > 3^\circ$ or $\epsilon_{\mat{t}'} > 3^\circ$, where
\begin{equation}
    \epsilon_{\mat{R}'} = (180 / \pi) \arccos \left( \left( \text{tr} \left( \mat{R}' \mat{R}^\text{T} \right) - 1 \right) / 2\right)
     \label{eq:rotation-error}
\end{equation}
is the rotation error and
\begin{equation}
    \epsilon_{\mat{t}'} = (180 / \pi) \arccos (\mat t^\text{T} \mat t')(|\mat t| |\mat t'|)
     \label{eq:translation-rotation-error}
\end{equation}
is the angular translation error in degrees~\cite{IMC2020}.
This ensures that the homography is consistent with the scene geometry and it can be recovered from the correspondences. 

Finally, we keep only a single estimatable homography for each test case since the purpose of the benchmark is to compare robust estimators, \eg RANSAC, that find only a single model.
Thus, an image pair with $k$ homographies is split into $k$ test scenes.
Each of them is generated by removing the inliers of the other estimatable homographies. 
Note that we keep those correspondences that are shared between the current homography and any other one.

\begin{figure}[t]
  	\centering
    \includegraphics[width=0.36\columnwidth,angle=90]{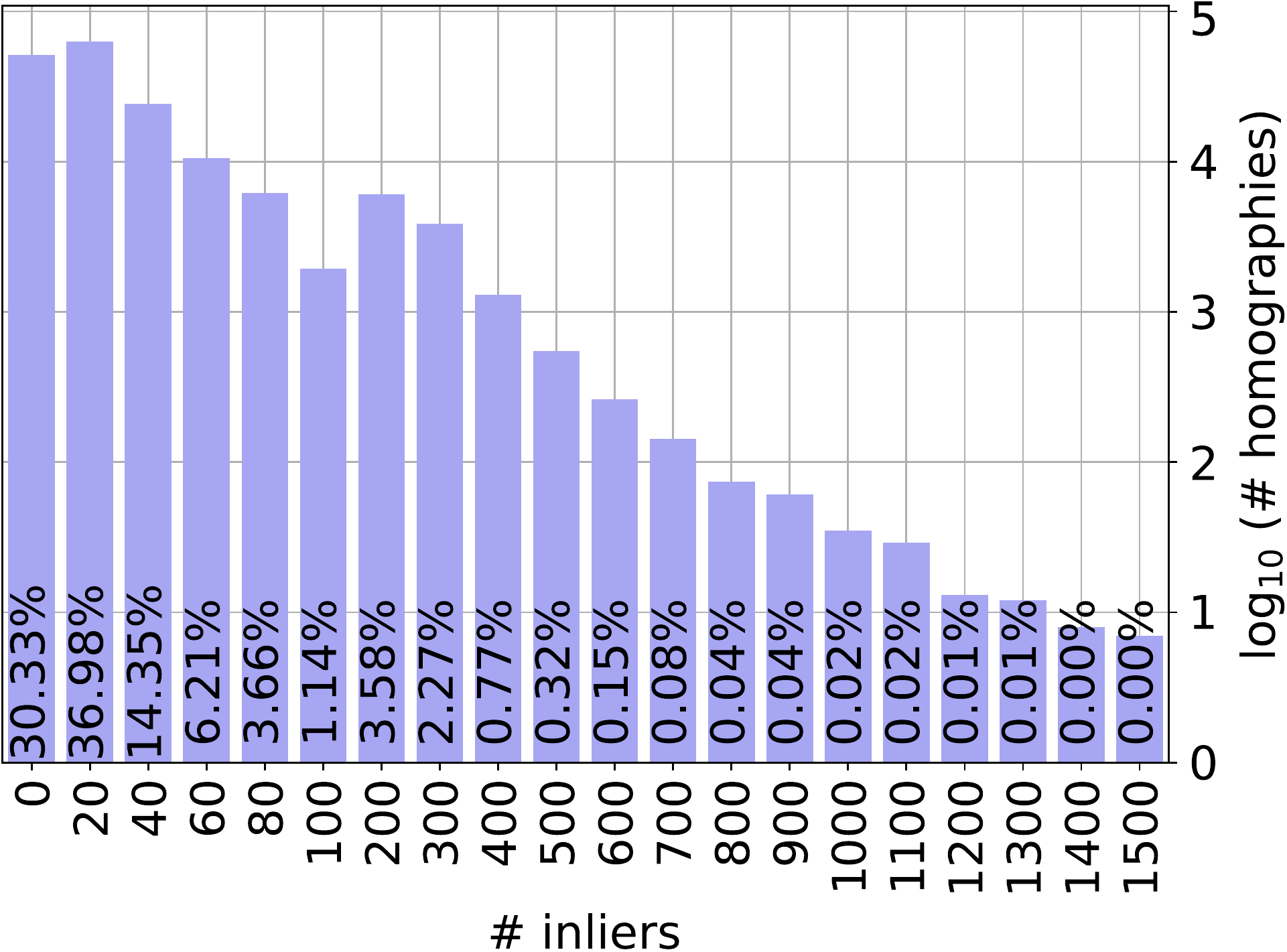}
    \includegraphics[width=0.36\columnwidth,angle=90]{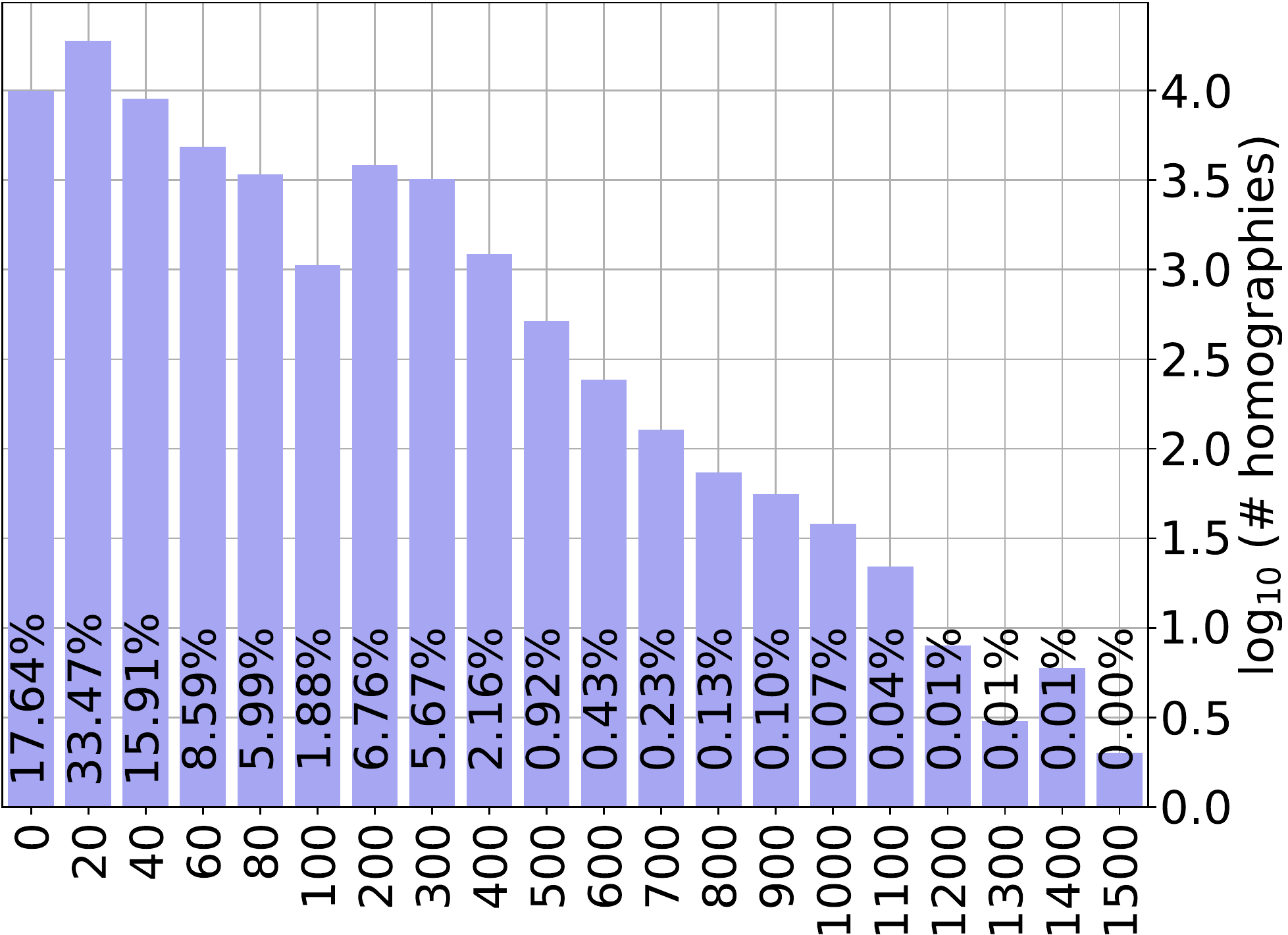}
	\caption{ 
	   Inlier number distribution in the training (left) and test (right) set of the HEB dataset. 
	}
    \label{fig:visualization_2}
    \vspace{-0.5em}
\end{figure}

\section{Homography Evaluation Benchmark}
The tentative correspondences are obtained from the mutually nearest RootSIFT matches minus the inliers of the other planes in the image pairs. 
The full input information, available to the methods is a set of $N$ correspondences $\{C_i\}_{i=1}^N$, each consisting of
 ${(x_i, y_i, \phi_i, s_i, x'_i, y'_i, \phi'_i, s'_i, \text{SNN ratio})}$, where $x_i$, $y_i \in \mathbb{R}$ are the point coordinates, $\phi \in [0, 2\pi)$ is the SIFT feature orientation, $s \in \mathbb{R}$ is the scale, and SNN ratio is Lowe ratio~\cite{SIFT2004} and $'$ denotes the second image.

The dataset is split into two disjoint parts. The training set contains two scenes -- Alamo and NYC Library. 
The test set contains the remaining nine scenes. 
While the training set might not be large enough to allow training models from scratch, it allows to set the parameters of models and traditional algorithms, such as inlier-outlier threshold. 

In Figures~\ref{fig:visualization_1} and \ref{fig:visualization_2}, properties of the HEB dataset are visualized. 
%
%
The left plots of Fig.~\ref{fig:visualization_1} report the $\log_{10}$ number of homographies (vertical axis) having a particular inlier ratio (horizontal).
The figures clearly demonstrate that the benchmark is extremely challenging since approximately the $80\%$ of the homographies in the dataset have at most $0.1$ inlier ratio. 
The training set shows similar statistics with marginally fewer cases with high inlier ratio.

The plot in the right of Fig.~\ref{fig:visualization_1} shows histograms of the angle between the translations $\textbf{t}$ and plane normals $\textbf{n}$.
The $0^\circ$ case can be interpreted as a camera moving backwards from the plane.
When the angle is $90^\circ$, the camera moves sideways.
At $180^\circ$, the camera moves towards the observed plane. 
It can be seen that all possible directions are well-covered both in the test and training sets.

\begin{figure*}[t!]
  	\centering
 	\caption{Comparison of $\mathbf H$ quality metrics.
 	Results averaged over all datasets: (a), (b) average median number of inliers versus mAA of the pose error and mAA of the pixel re-projection error. While more inliers often imply better accuracy, it is not always the case, and methods may have different accuracy with similar numbers of inliers.
	In plots (c), (d), the mAA of the pose versus mAA of the re-projection errors and the mAA of rotation-only component (used in IMC~\cite{IMC2020}) are shown. LMEDS and LSQ are omitted here. The re-projection error is a good proxy for pose accuracy with two exceptions. (d) While the pose accuracy and scale-less rotation-only mAA~\cite{IMC2020} are well-correlated, the method ranking is significantly affected by the metric.}
    \includegraphics[width=0.22\linewidth]{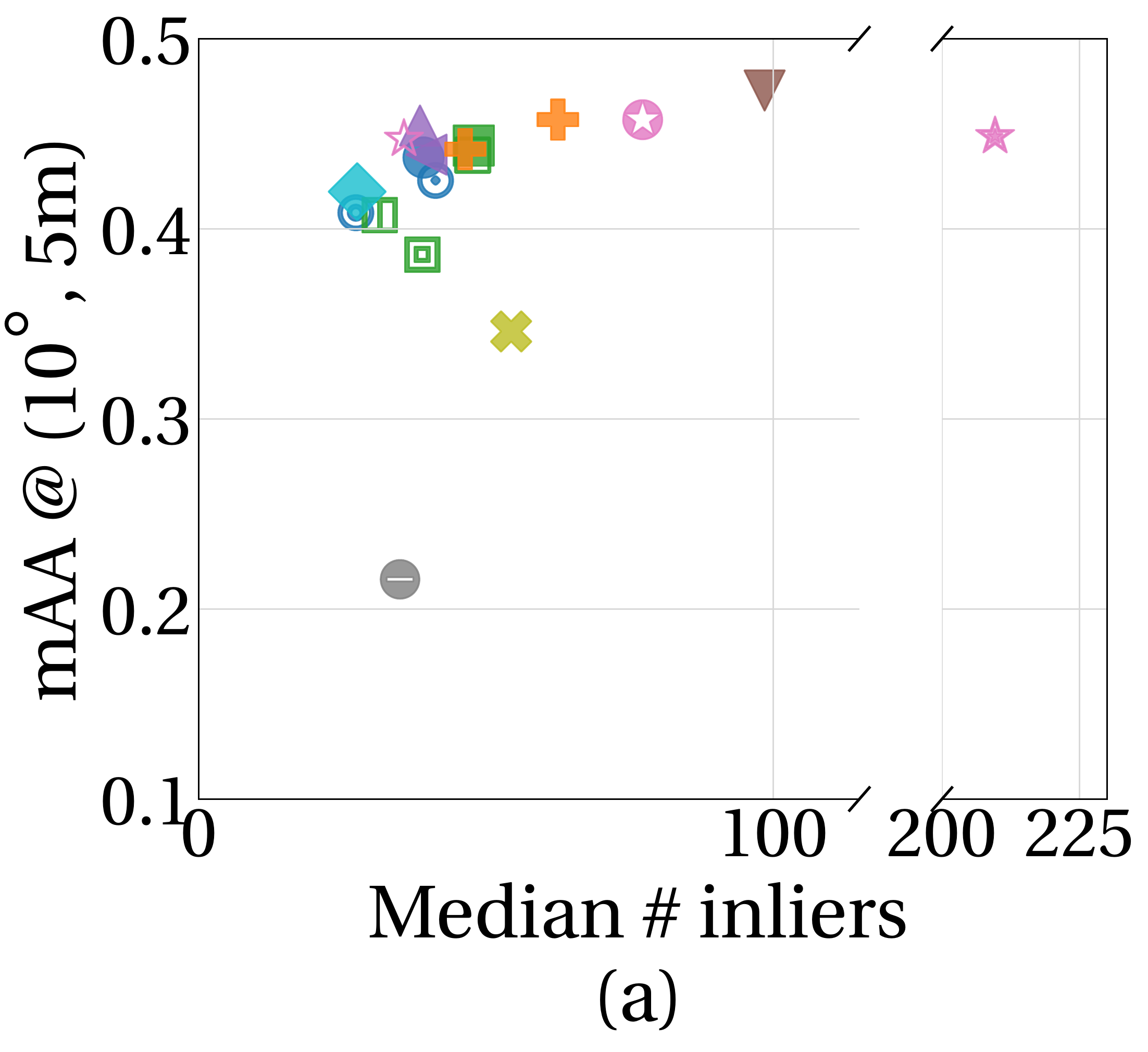}
    \includegraphics[width=0.22\linewidth]{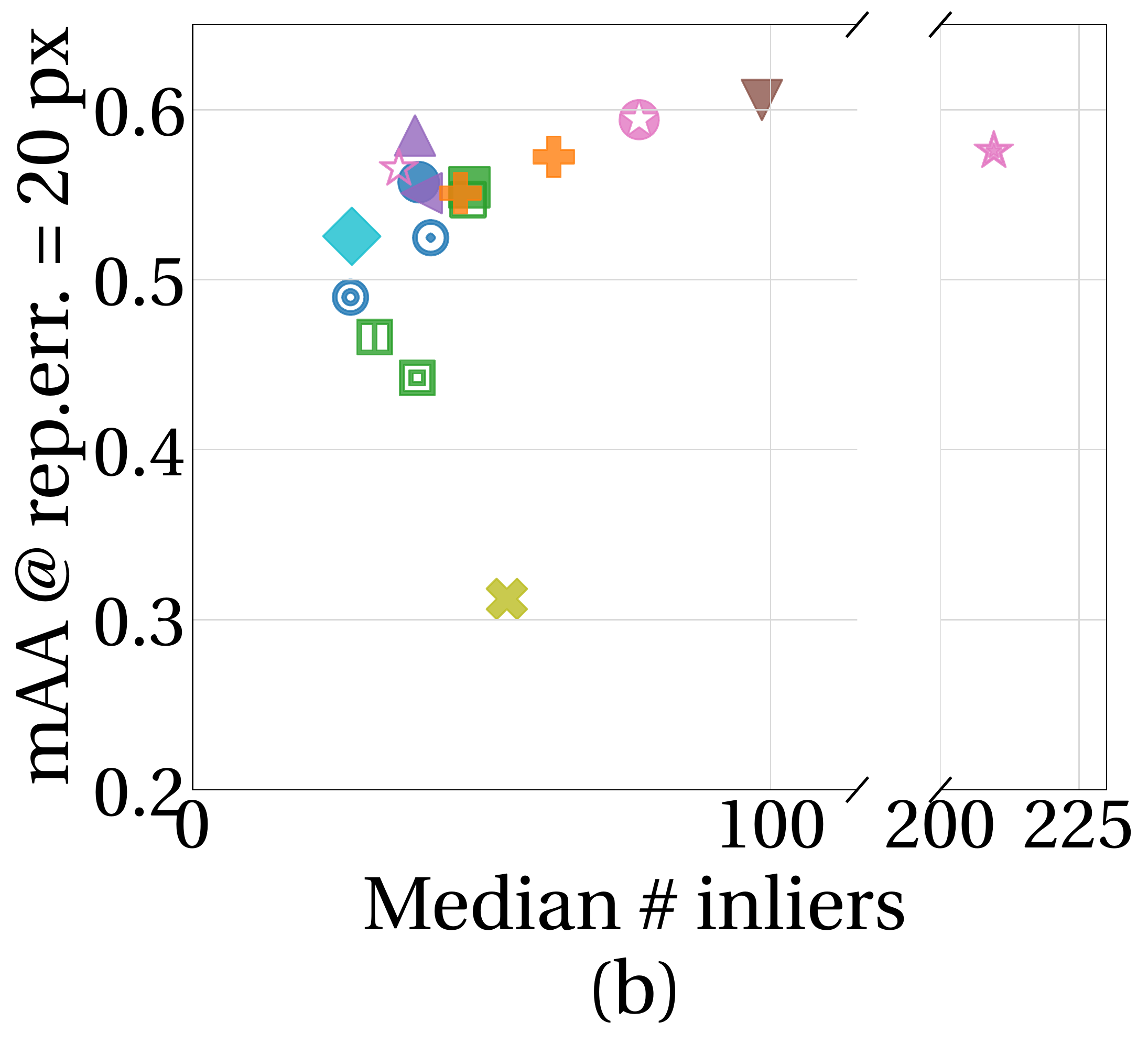}
    \includegraphics[width=0.195\linewidth]{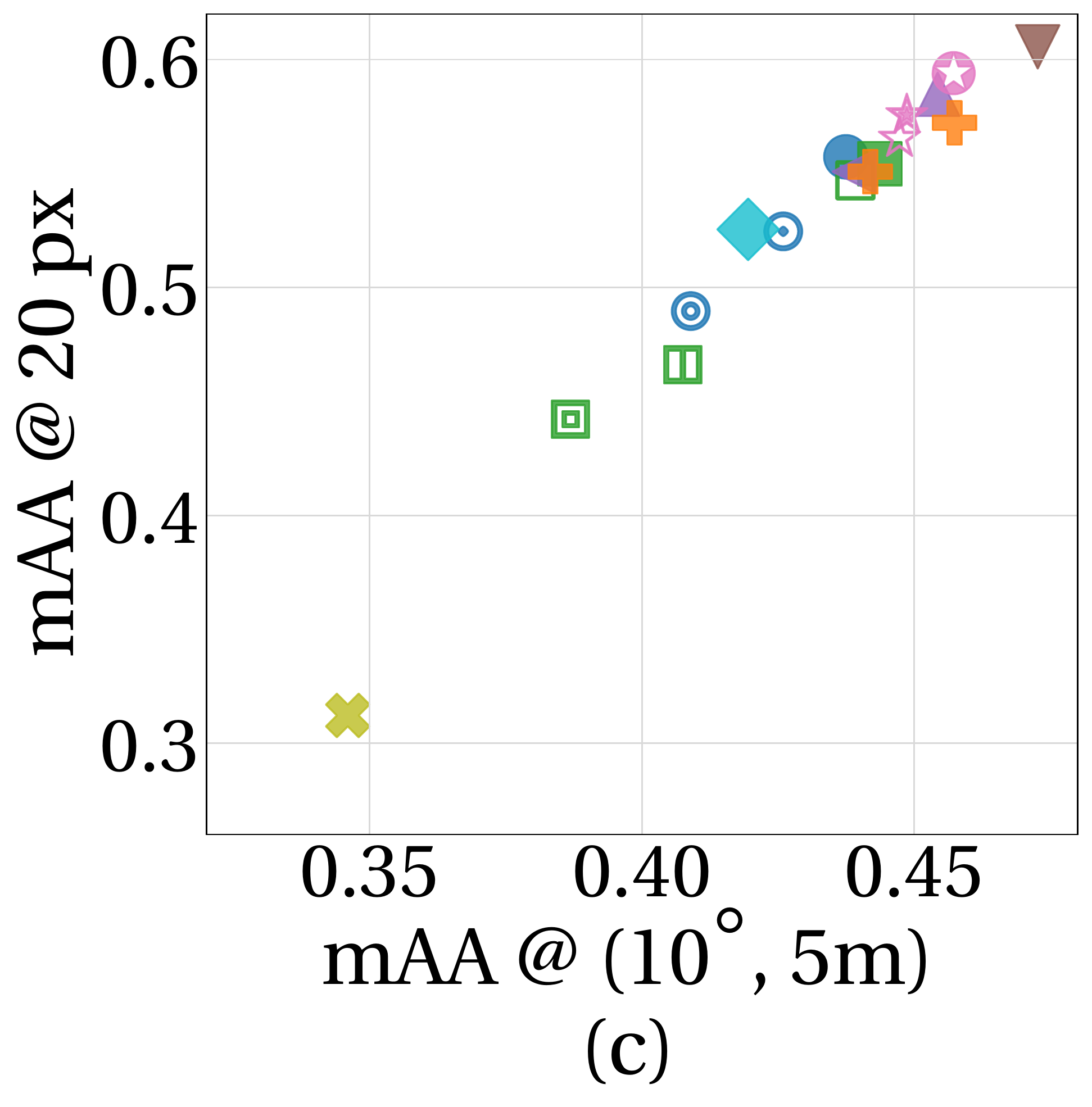}
    \includegraphics[width=0.22\linewidth]{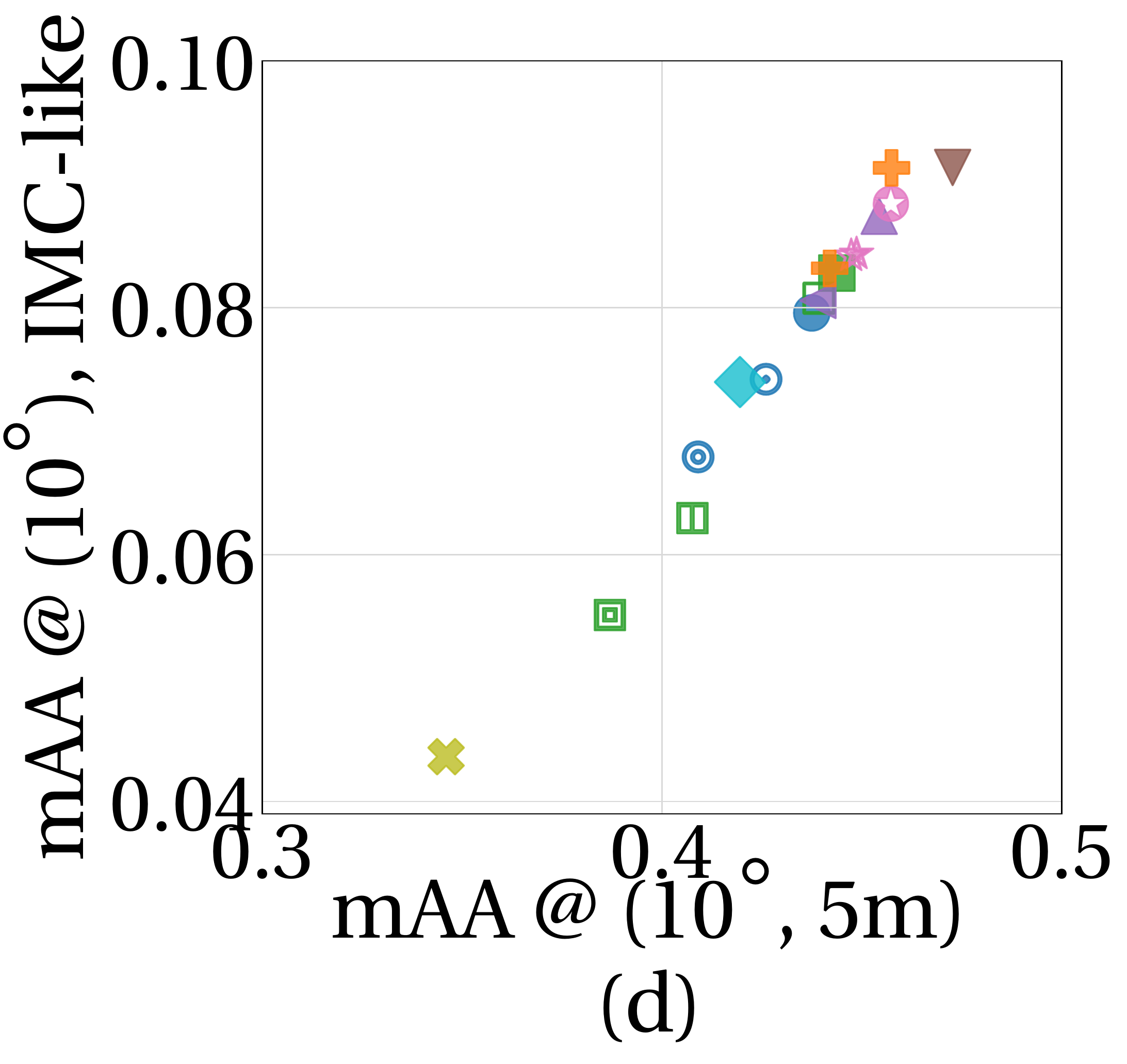}
    \label{fig:metrics-comparison}
     \vspace{-0.5em}
\end{figure*}
\begin{figure*}[t!]
  	\centering
  	\fbox{
  	  \includegraphics[width=1.0\linewidth]{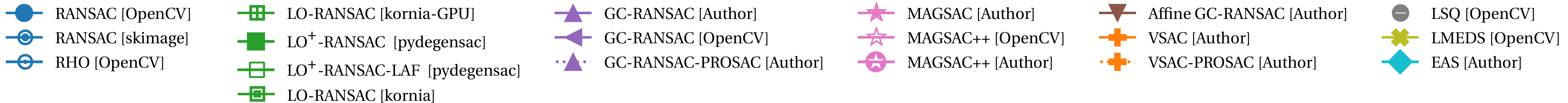}}
  	  \\[2ex]
        \includegraphics[width=0.35\linewidth]{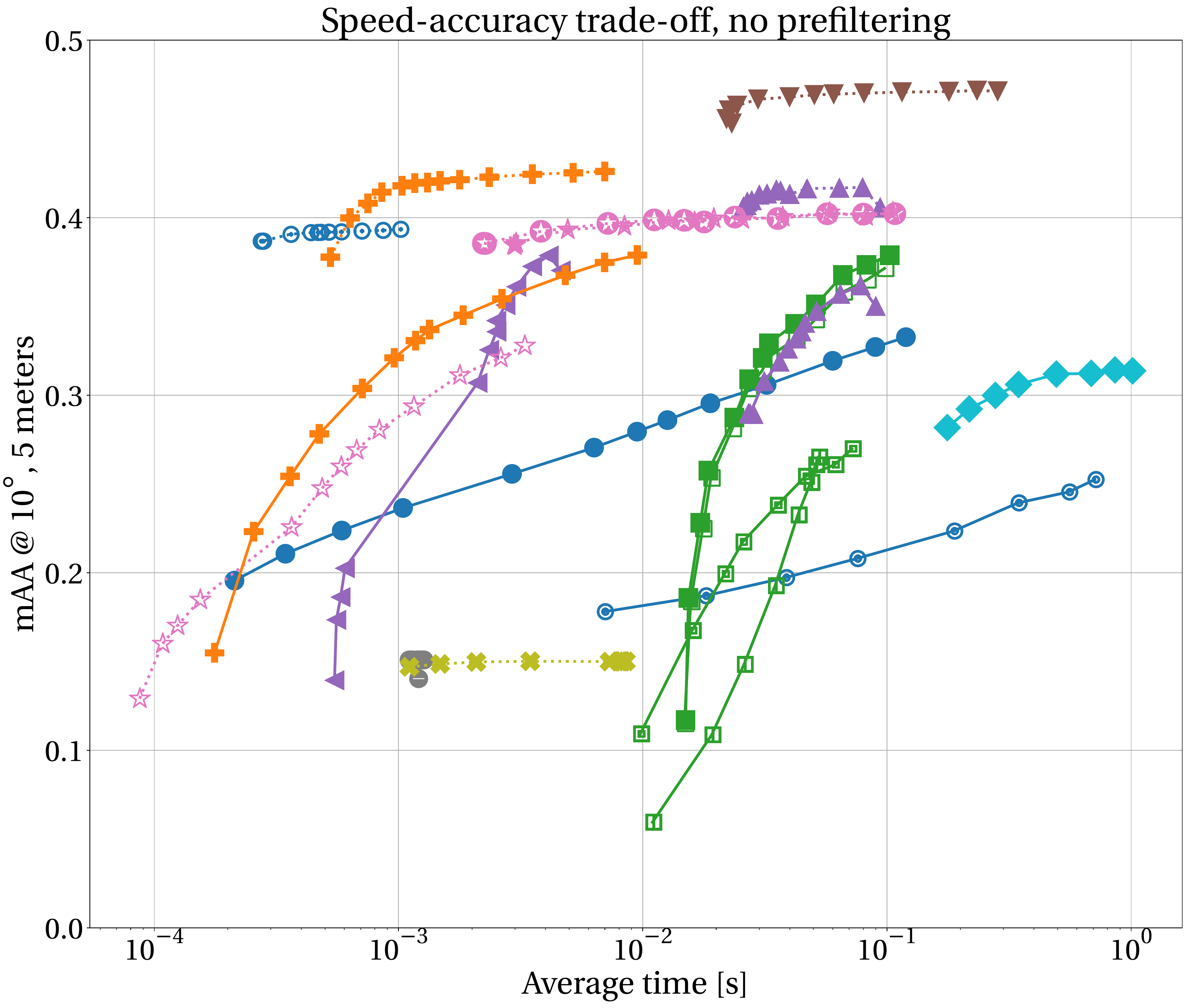}
        \includegraphics[width=0.35\linewidth]{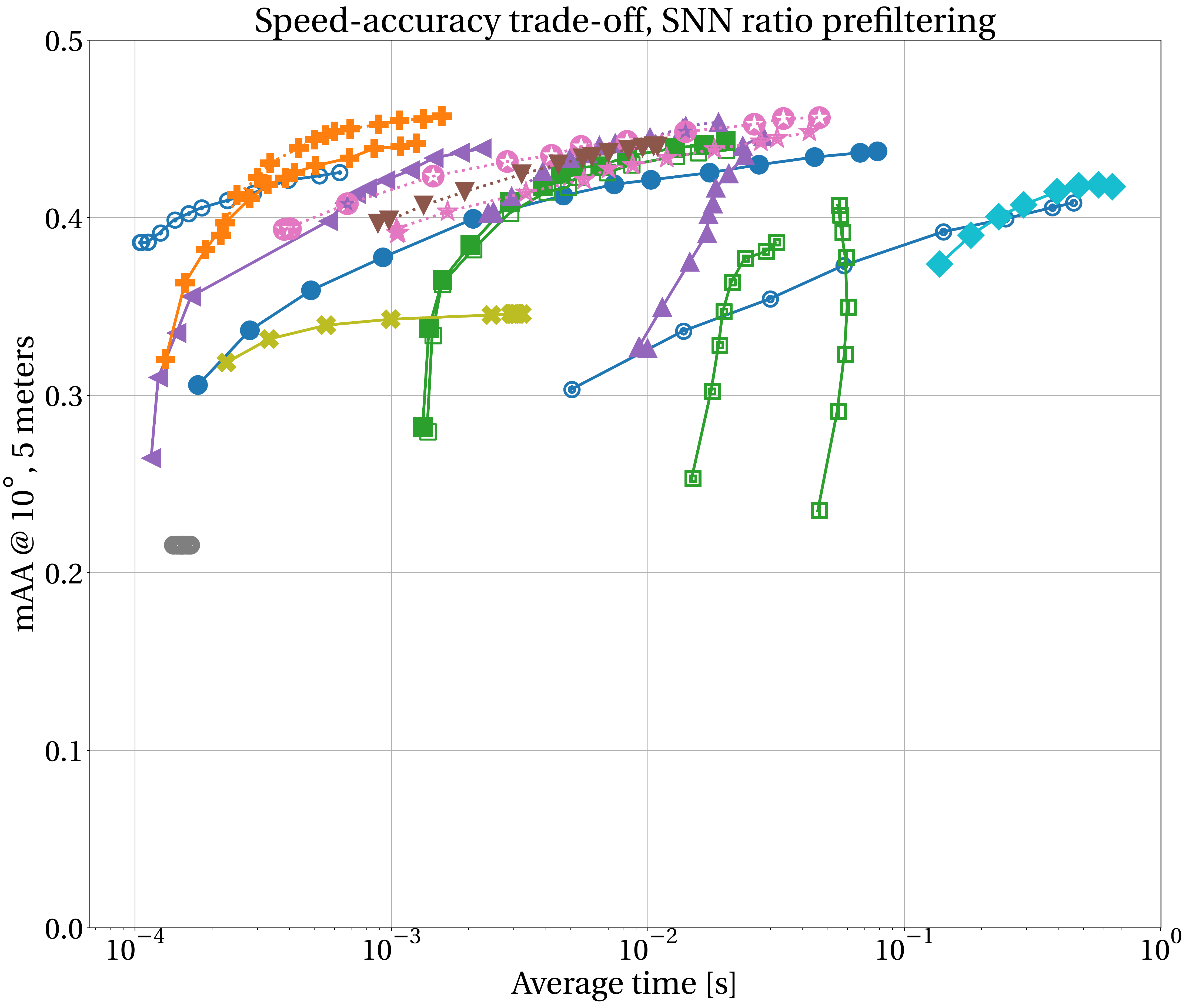}
	\caption{Speed-accuracy comparison of homography estimators on HEB test set, average over images and scenes.
	The max.\ number of iterations was varied from $10^1$ to $10^3$. Note the logarithmic scale of the time axis.
	Left -- no prefiltering except mutual nearest neighbor check.
	Right -- mutual nearest neighbor and SNN (Lowe) ratio check. PROSAC sorting is indicated by a dashed line.}
	 \vspace{-0.5em}
    \label{fig:traditional-methods}
\end{figure*}


In Fig.~\ref{fig:visualization_2}, the inlier numbers are shown. 
In $30\%$ of the cases, the homographies have fewer than $20$ inliers, making the robust estimation challenging, especially when the outlier number is high. 
It is important to note that the success, in practice, depends more on the inlier number than the inlier ratio. 
This is caused by the fact the outliers often tend to form spatially coherent structures misleading the estimator if the inliers are sparsely distributed in the scene~\cite{ivashechkin2021vsac}. 
The majority of the homographies have fewer than $50$ inliers. The same distribution holds for the training set. 

\section{Experimental Protocol}

Our evaluation protocol is largely influenced by the Image Matching Benchmark~\cite{IMC2020}. However, we made several important changes, described below.

\noindent
\textbf{Metrics.}
We compute a range of per-pair metrics from one of the following three groups.

\noindent\textit{(i) Pose-based:} Eqs.~\eqref{eq:rotation-error}, \eqref{eq:translation-rotation-error} and absolute translation error: 
\begin{equation}
    \epsilon_{\mat{t}_{abs}'} ={|\mat t - \mat t'|_2}.
    \label{eq:abs-translation-error}
\end{equation}
\noindent\textit{(ii) Ground truth correspondences-based:} re-projection error of the GT correspondences with estimated homography:
\begin{equation}
    \epsilon_{\text{repr}} ={|\mat x - {\cal H}( \mat x')|_2}.
    \label{eq:re-projection-error}
\end{equation}
the homography operator $\cal H$ transforming the non-homogeneous image coordinates $\mat x'$.

\noindent\textit{(iii) Self-supervised:} number of inliers, run-time.

The per-homography metrics are accumulated into scene-metrics by the  (a) mean, (b) median and (c) calculating mean average accuracy (mAA) with thresholds:
from 1$^\circ$ to 10$^\circ$ for angular metrics, from 0.1 m to 5 meters for absolute translation error Eq.~\eqref{eq:abs-translation-error} and from 1 to 20 pixels for re-projection error Eq.~\eqref{eq:re-projection-error}. 
The thresholds resemble the ones used in the visual localization literature~\cite{aachendaynight2}.

Since the scale can not be recovered from an essential matrix or homography~\cite{hartley2003multiple}, we assign the GT absolute scale to the estimated translation $\mat t'$. 
There is an important difference between measuring the absolute translation error and the purely angular one in Eq.~\eqref{eq:translation-rotation-error} as done in IMC~\cite{IMC2020}. 
When the baseline $\mat t$ is small, \eg, a few centimeters, the noise in the camera position has a large effect on the translation angle.
Thus, Eq.~\eqref{eq:translation-rotation-error} distorts the evaluation by returning large errors even when the camera barely moves in the real-world.
We select the averages of the rotation and translation mAA scores to be our main metric.

\noindent\textit{Metrics comparison.}
%
We plot the angular pose accuracy vs.\ metric pose accuracy in Fig.~\ref{fig:metrics-comparison} (right). 
They are mostly in agreement, except for a few methods, \eg, EAS~\cite{fan2021efficient} and Affine GC-RANSAC~\cite{barath2020making}.
The mAA of the re-proj.\ error is also in agreement with the mAA of the pose error (Fig.~\ref{fig:metrics-comparison}; $3$rd) with some exceptions, \eg, LO$^{+}$-RANSAC. 

The number of inliers (Fig.~\ref{fig:metrics-comparison}, two left graphs) greatly depends not only on image resolution,  but also on the inlier threshold and particulars of each algorithm -- MAGSAC outputs many more inliers, while having similar pose accuracy to other methods, while the LMEDS pose is much worse with the same number of inliers as the rest. 

\begin{figure*}[t]
  	\centering
        \framebox{
         \begin{minipage}{0.90\linewidth}
  	\hspace*{-0.03\linewidth}\includegraphics[width=0.98\linewidth]{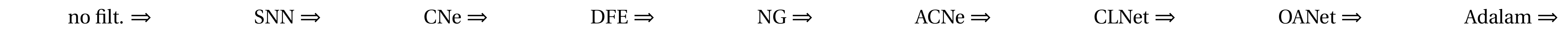}
        \includegraphics[width=0.98\linewidth]{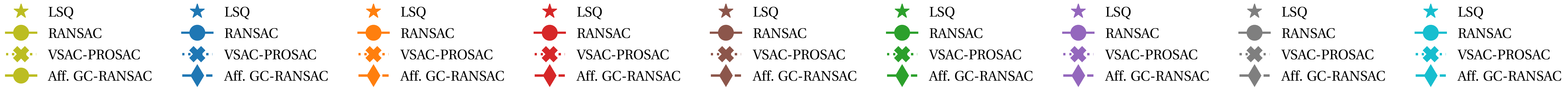}
        \end{minipage}
        }\\[2mm]
        \includegraphics[width=0.35\linewidth]{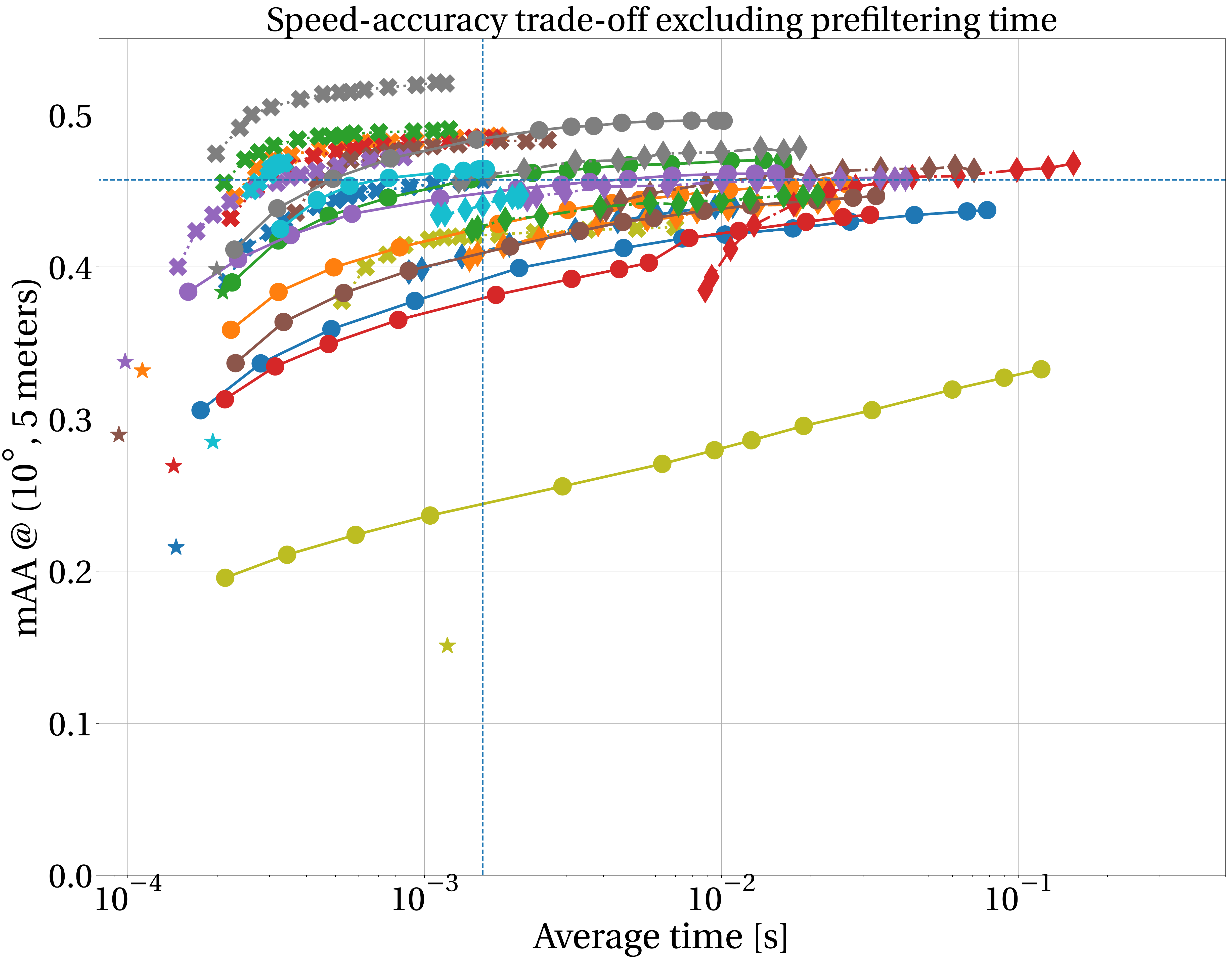}
        \includegraphics[width=0.35\linewidth]{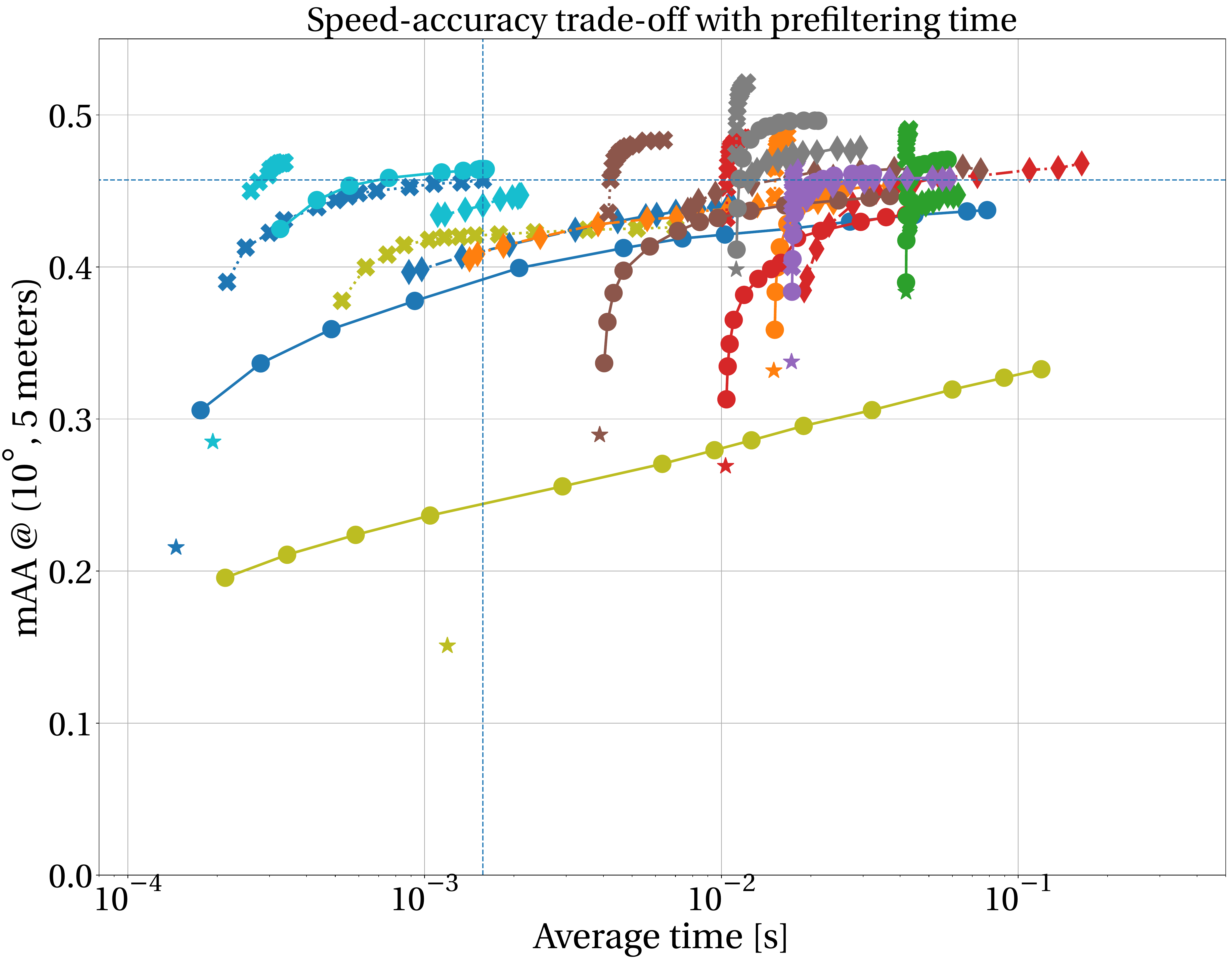}

	\caption{Speed-accuracy comparison of the classical SNN ratio and deep prefiltering on OpenCV LSQ and RANSAC algorithms, Affine GC-RANSAC~\cite{barath2020making} (best accuracy w/o prefiltering) and VSAC-PROSAC~\cite{ivashechkin2021vsac} (best accuracy with SNN prefiltering). 
	Left -- RANSAC time ($\log_{10}$ scale), right -- time of RANSAC and the deep prefiltering.
	The best result of SNN ratio is marked by blue dashed lines. 
	}
    \label{fig:deep-methods}
\end{figure*}


\noindent
\textbf{Training and Test Protocols.}
One of the drawbacks of the existing homography estimation datasets is the lack of tuning and test protocols. 
We propose the following procedure for fair evaluation. 
The main principle is as follows: one should not not make more than one or two evaluation runs on the test set. 
That it why all the hyper-parameters of the algorithms are fixed when running on the test set.
The tuning and learning are done on the training set, which has similar, but not equal properties and no overlap in terms of content with the test set. 
We tune all the hyper-parameters with grid search for simplicity.

\textit{Training protocol.} 
We fix number of iterations to 1000 for all methods. 
With each method, grid search is performed on the training set to determine the optimal combination of the hyper-parameters, such as inlier-outlier threshold $\theta$, the SNN ratio threshold and other algorithm-specific parameters, such as the spatial weight of GC-RANSAC. 
Note that, unlike IMC~\cite{IMC2020}, inlier-outlier and SNN thresholds are tuned jointly and not consequently -- we found that it leads to slightly better hyper-parameters.

We tested the robust estimators on correspondences filtered by the predicted score of recent deep learning models.
After obtaining the scores, we post-processed them in one of the two ways: (a) thresholding the scores at $\theta$ and removing tentative correspondences below it; and (b) sorting the correspondences by their score and keeping the top $K$ best. 
Both $\theta$ and $K$ were found by running grid search on the training set similarly as for other hyper-parameters.


\textit{Test protocol.}
After fixing all hyper-parameters, we run the algorithms on the test set, varying their maximum number of iterations from \num{10} to \num{10000} (to \num{1000} for methods significantly slower than the rest, \ie, scikit-image RANSAC, EAS and kornia-CPU) to obtain a time-accuracy plot. 
The~algorithm terminates after its iteration number reaches the maximum. 
Note that, unlike in IMC~\cite{IMC2020}, such experiments are performed on the test, not training set. 
 
 


\noindent
\textbf{Methods for Homography Estimation.}
We give a brief overview of algorithms that we compare on HEB.
Note that we consider it important to compare not just the algorithms as published in their respective papers but, also, their available implementations. 
Even though it might seem unfair to compare a method implemented in Python to C++ codes, the main objective is to provide useful guidelines for users on which algorithms and implementations to use in practice. 

\textit{Traditional Algorithms.}
In all tested methods, the normalized DLT algorithm runs both on minimal and non-minimal samples. 
We found that the implementation is as important as the algorithm itself, thus, we define a method by its name and the library in which it is implemented. 

We compare the OpenCV implementations of RANSAC~\cite{fischler1981random}, LMEDS~\cite{rousseeuw1984least}, LSQ, RHO~\cite{bazargani2018fast}, MAGSAC++~\cite{barath2020magsacpp}, and Graph-Cut RANSAC~\cite{barath2018graph}. 
The RANSAC implementation as in the scikit-image library~\cite{scikit-image}.
Unlike OpenCV RANSAC, which is implemented in optimized C++ code, scikit-image is implemented in pure Python with the help of numpy~\cite{numpy}.
LO-RANSAC~\cite{LORANSAC2003} as implemented in the PyTorch~\cite{pytorch}-based kornia library~\cite{kornia2019}. 
LO-RANSAC$^+$~\cite{fixingLORANSAC2012} implemented in the pydegensac library with and without local affine frame (LAF) check~\cite{Mishkin2015MODS}. 
The Graph-Cut RANSAC, MAGSAC~\cite{barath2021marginalizing}, MAGSAC++ and VSAC~\cite{ivashechkin2021vsac} algorithms implemented by the authors. 
While MAGSAC and MAGSAC++ uses the PROSAC sampler~\cite{chum2005matching} as default, we run GC-RANSAC and VSAC with and without PROSAC.
We also evaluate the deterministic EAS algorithm~\cite{fan2021efficient} provided by the authors.
EAS is implemented in pure Python using the numpy~\cite{numpy} package. 

Also, we apply the affine correspondence-based GC-RANSAC~\cite{barath2020making} with its implementation provided by the authors. 
Since our benchmark does not have affine correspondences, we approximate them using SIFT features. 
Given rotations $\alpha_1, \alpha_2 \in [0, 2\pi]$ and scales $s_1, s_2$ in the two images for a correspondence, the affine transformation is calculated as $\textbf{A} = \textbf{J}_2 \textbf{J}_1^{-1}$, where $\textbf{J}_i = \textbf{R}_i \textbf{S}_i$, matrix $\textbf{R}_i$ is the 2D rotation by $\alpha_i$ degrees, and $\textbf{S}_i$ is the 2D scale matrix uniformly scaling by $s_i$ along the axes, $i \in [1, 2]$.

%

\textit{Deep prefiltering.}
The standard two-view matching pipeline with SIFT or other local features uses the SNN test~\cite{SIFT2004} to filter out unreliable matches before running RANSAC~\cite{IMC2020,brachmann2019ngransac,Efe_2021_ICCV}. 
Recently, it was shown~\cite{cne2018,dfe2018} that using a neural network for correspondence prefiltering might provide benefits over the SNN ratio test.

We evaluated how using models~\cite{cne2018,dfe2018,brachmann2019ngransac,oanet2019,acne2020,clnet2021} for correspondence prefiltering for uncalibrated epipolar geometry help in homography estimation. 
For our study, we took pre-trained models, provided by the authors of each paper and use them for scoring the tentative correspondences. 
We emphasize that we neither trained, nor fine-tuned them for the homography estimation task, so their performance is sub-optimal.
The reason why we did not take the pre-trained models for homographies is that authors do not provide them.
Unless stated otherwise, all the pre-trained models we used, were trained on a subset~\cite{cne2018} of YCC100M dataset for fundamental matrix estimation.


\section{Experiments}

\noindent
\textbf{Traditional Methods.}
The pose errors are shown in Fig.~\ref{fig:traditional-methods}.

\textit{No-prefiltering.} 
This is the setup, where the difference between methods is the most pronounced. 
The most accurate method in all metrics is Affine GC-RANSAC that exploits the orientation and scale of SIFT features and, thus, reduce the combinatorial complexity of the problem. 
The second most important feature is PROSAC sampling, which improves the results of VSAC and GC-RANSAC by up to 10 percentage points. 
The optimized implementation matters a lot in terms of speed -- python-based skimage RANSAC, EAS and kornia-CPU are up to 1-3 orders of magnitude slower than the other RANSACs. 
Kornia-GPU is on par in terms of speed with OpenCV RANSAC or pydegensac LO$^+$-RANSAC, but is worse in terms of accuracy. 
Even with the same language (C++), the speed and even the accuracy of different implementations of GC-RANSAC and MAGSAC++ vary significantly. 

\textit{Prefiltering with SNN ratio}.
With optimal SNN ratio filtering, the difference between methods becomes smaller and most of the advanced RANSACs show similar accuracy, \eg, LO$^+$ and GC-RANSAC. 
For most methods, the best SNN threshold is 0.6, which is stricter than the widely used 0.8. 
We believe that it is due to HEB having small inlier ratios, hence requiring aggressive filtering. 
%
The RHO algorithm is still the leader in top-speed part, outperformed by VSAC-PROSAC with increasing time budget. 
Affine GC-RANSAC is the one which benefits from correspondence prefiltering the least, both in terms of speed and accuracy. 

As expected, LSQ fitting and LMEDS yield inaccurate results in all cases due to the high inlier ratio in the dataset.  
Interestingly, the recently proposed EAS algorithm~\cite{fan2021efficient} leads to highly inaccurate results both in the SNN-filtered and unfiltered cases. 
It is also surprising that affine GC-RANSAC~\cite{barath2020making} with using approximated affine correspondences only (from the SIFT orientations and scales) is the top-performing method in the unfiltered case and is among the best ones when SNN filtering is applied.
This highlights the importance of using higher-order features to reduce the sample size in RANSAC.
Due to the small sample size, the combinatorics of the problem is reduced, thus improving randomized RANSAC-like robust estimation. 




\noindent
\textbf{Deep prefiltering.}
Results are shown in Fig.~\ref{fig:deep-methods}. 
The top row shows the combined pose error, while the bottom one shows the errors either in the rotation or in the translation. 
The best deep prefiltering methods provide an accuracy boost to advanced RANSACs of the similar magnitude, as switching from the no-filtering to SNN ratio filtering. 
However, not all methods are equal: there is a clear distinction between earlier methods like DFE, CNE and NG, and later models like OANet, ACNe and CLNet. 
The latter ones use specialized architectures, while DFE, CNE and NG are based on batch-normalized MLPs.
OANet provides the best results, it is also the only model among the leaders which uses side information -- SNN ratio -- as an input. It is also interesting that the vanilla OpenCV RANSAC with OANet or CLNet prefiltering performs similarly to VSAC + SNN ratio in terms of accuracy.
LSQ with deep filtering performs similarly to RANSAC with SNN-ratio filtering and better than RANSAC without prefiltering at all. 


Finally, we show the time-accuracy plot in Fig.~\ref{fig:deep-methods} (right) when the deep prefiltering (on NVIDIA V100 GPU) time is taken into account. 
It is at least 5-10 ms per image pair for the fastest methods (NG, DFE and CLNet), which potentially is a limitation for real-time applications, especially when running on a smart device without GPU.  

\paragraph{An application: uncertainty of SIFT keypoints.}
The uncertainty of popular detectors and their implementations is unknown or incomparable, \eg, only refer to a certain resolution. 
Our goal is to determine bias and variance of angular, scale, and positional transformations of detected correspondences of SIFT keypoints $\{C_i\}_{i=1}^N$ and -- if possible -- compare it to previous results. This may be a motivation to use the scaled rotation as an approximation for the local affine transformation. 

The positional uncertainty of SIFT keypoints is known to be approximately 1/3 pixel (see \cite{forstner2016photogrammetric} p.681, \cite{laebe*08:quality} Tab.6). 
The standard deviations (STD) of the keypoints depend on the detector scales (see \cite{forstner2016photogrammetric} p.681, \cite{zeisl*09:estimation} Eq.(15)). 
We are not aware of investigations into the uncertainty of the directions and scales.
The  SIFT detector (in OpenCV) uses an orientation histogram with 36 bins 
of 10 degrees. 
Assuming an average STD of less than three times the rounding error $10^\circ/\sqrt{12} \approx 2.89^\circ$, the average STD of $\alpha_i=\phi'_i-\phi_i$ is approx. 12$^\circ$, the factor three taking care of other model errors. This large uncertainty may be useful in cases where the rotation between keypoints is large. 

While the reference scale ratios easily can be determined from a local reference affinity $\widetilde{\mat A}_i$, derived from $\widetilde{\mat H}_i$, the reference rotations $\widetilde{\alpha}_i$ requires care. There are two approaches to obtain reference rotations: (1) comparing direction vectors $\mat d(\phi'_{i})$ in the second image with the transformed direction $\mat d(\phi_{i})$ in the first image, and (2) deriving a local rotation from the reference affinity matrix $\widetilde{\mat A}_i$ and compare it to $\alpha_i$. 

We apply following approach: 
approximate the projective transformation by a local affinity $\widetilde{\mat A}_i \in \mathbb{R}^{2 \times 2}$, and, 
decompose $\widetilde{\mat A}_i$ into reference scale ratio $\widetilde{r}_i$, rotation angle $\widetilde{\alpha}_i$, and two shears $\widetilde{\mat p}_i\in \mathbb R^2$. We investigated QR, SVD and an exponential decompositions, namely decomposing the exponent $\widetilde{\mat B_i}$ of  $ \widetilde{\mat A_i}=\exp(\widetilde{\mat B_i})$ additively (see
supplement). We evaluate the differences $\Delta\alpha_i = \widetilde{\alpha}_i - \alpha_i$ between observed and reference angles. The bias $\mathbb E(\Delta\alpha_i)$, \ie the mean of $\Delta\alpha_i$ and the STD $\sigma_{\Delta\alpha_i}=\sqrt{\mathbb D(\Delta\alpha_i)}$ of the rotation differences $\Delta \alpha$, for the OpenCV SIFT detector empirically lead to an estimated STD  of the rotation $\hat{\sigma}_{\Delta\alpha_i} =11.8^\circ$, which is close to the above mentioned expectation. 

Each of the three approaches leads to different reference rotations $\widetilde{\alpha}_i$. Rotation $\widetilde{\alpha}_i$ is effected by the shears $\widetilde{\mat p}_i$ in $\widetilde{\mat A}_i$. If the shears are small, all three methods yield similar rotations. The magnitude $|\widetilde{\mat p}_i|^2$ of the shears can be approximated by the condition number $\mbox{cond}(\widetilde{\mat A}_i)$. 
To evaluate the rotations $\alpha_i$ of the keypoint pairs, we restrict the samples to those with condition number $< 1.5$, which for image pairs in normal pose roughly is equivalent to slopes of the scene plane below 25$^\circ$.
Moreover, we show the comparison of angular residuals between $\mat d'_i = [\cos(\phi_i') \; \sin(\phi_i')]^\text{T}]$ and the one obtained by affinely transformed $\mat d_i = [\cos(\phi_i)\; \sin(\phi_i)]^\text{T}]$, \ie with $\widetilde{\mat A}_i \mat d_i$. The average deviations are similar to those obtained with the decomposition methods, see the details in the suppl. material. 

The scale ratio $r_i = s_i' / s_i$ of a keypoint pair and its ratio $\Delta r_i = r_i / \widetilde{r}_i$  to the reference ratio $\widetilde{r_i}$ should lead to $\mathbb E(\Delta r_i) = 1$. Further, we use a weighted log-ratio, measured as $\rho_i = \log(\Delta r_i) / {\widetilde{r}_i}$ which should follow $\mathbb E(\rho_i)= 0$, and takes into account the intuition, that larger scales are less accurate. The OpenCV implementation of the SIFT detector empirically leads to $\hat{\sigma}_{\rho_i}=0.51$ (see the suppl. material). 
Obviously, the scales from the detector may on average deviate by a factor $1.6 \approx \exp(0.51)$ in both directions. 

The positional residual of each keypoint pair is characterized by the squared mean reprojection error
$\epsilon_{x_i} = \sqrt{(
|\mat x'_i-{\widetilde{\cal H}}(\mat x_i)|_2^2+|\mat x_i-\widetilde{{\cal H}}^{-1}(\mat x_i')|_2^2)/8}$,
the factor 8 guaranteeing that $\epsilon_{x_i}$ can be compared to the expected uncertainty of the coordinates. For the OpenCV SIFT detector, we empirically obtain a positional uncertainty of $\epsilon_{x_i}$ as $\hat{\sigma}_{x} \approx 0.67$ pixels. The STD is a factor two larger, than expected, which might result from accepting small outliers.

%
%

\section{Conclusion}

A large-scale dataset containing roughly 1000 planes (Pi3D) in reconstructions of landmarks, and a homography estimation benchmark (HEB) is presented. 
The applications of the Pi3D and HEB datasets are diverse, \eg, training or evaluating monocular depth, surface normal estimation and image matching algorithms.  
As one possible application, we performed a rigorous evaluation of a wide range of robust estimators and deep learning-based correspondence filtering methods, establishing the current state-of-the-art in robust homography estimation. 
The top accuracy is achieved by combining VSAC~\cite{ivashechkin2021vsac} with OANet~\cite{oanet2019}.
In the GPU-less case, a viable option is to use VSAC~\cite{ivashechkin2021vsac}, OpenCV RHO~\cite{bazargani2018fast} or Affine GC-RANSAC with SNN test, depending on time budget. 
We also show that PROSAC -- a well known, but often ignored sampling scheme accelerates RANSAC by an order of magnitude. 
Exploiting the SIFT orientation and scale has clear benefits in Affine GC-RANSAC and it can be used in other approaches as well, \eg, VSAC.
%
The \textit{whole} dataset, including the reconstruction with absolute scale, and the tools for adding new features will be made available.

As another application, we show that having a large number of homographies allows for analyzing the noise in partially or fully affine-covariant features. 
As an example, we evaluate DoG features.
To the best of our knowledge, we are the first ones to investigate the actual noise in the orientation and scaling components of such features.

\paragraph{Acknowledgedment.} The work was funded by EU H2020 ARtwin No. 856994, EU H2020 SPRING No. 871245. 

%

\appendix
\section{Methods in the Main Experiments}

In this section, we describe the components of each algorithm compared in the main paper. 

\subsection{Traditional Algorithms}

In all tested methods, the normalized direct linear transformation~\cite{hartley2003multiple} (DLT) algorithm runs both on minimal and non-minimal samples to estimate homographies. 
The compared methods and implementations are the following.

\noindent \textbf{RANSAC (OpenCV).} The OpenCV implementation contains the following components in addition to the original RANSAC~\cite{fischler1981random} algorithm.
\begin{enumerate}[topsep=2mm, partopsep=1pt,itemsep=2pt,parsep=0pt]
    \item Sample cheirality check to reject minimal samples early if the implied plane flips between the two views.
    \item Levenberg-Marquardt numerical optimization minimizes the re-projection error on the final set of inliers.
    \item Single-sided re-projection error, measured in the second image, is used as point-to-model residual.
\end{enumerate}\vspace{2mm}

\noindent \textbf{LMEDS (OpenCV).} The OpenCV implementation of the Least Median of Squares algorithm~\cite{rousseeuw1984least} runs the same additional components as the OpenCV RANSAC. \vspace{2mm}

\noindent \textbf{LSQ (OpenCV).} The least-squares fitting by the normalized four-point algorithm implemented in OpenCV. \vspace{2mm}

\noindent \textbf{RANSAC (skimage).} The RANSAC as implemented in the scikit-image library~\cite{scikit-image}. It contains the following components in addition to the original RANSAC~\cite{fischler1981random} algorithm.
Single-sided re-projection error, measured in the second image, is used as point-to-model residual.
Unlike OpenCV RANSAC, which is implemented in optimized C++ code, scikit-image is implemented in pure Python with help of the numpy package~\cite{numpy}.\vspace{2mm}

\noindent \textbf{LO-RANSAC (kornia).} LO-RANSAC~
\cite{LORANSAC2003} as implemented in the kornia library~\cite{kornia2019}. It implements the LO-RANSAC as proposed in \cite{LORANSAC2003} (version 2 in Section 3), where the far-the-best model is obtained by running local optimization using all inliers.
Additional components:
\begin{enumerate}[topsep=2mm, partopsep=1pt,itemsep=2pt,parsep=0pt]
    \item Symmetric transfer error is used as point-to-model residual.
    \item Unlike LO-RANSAC, the kornia library uses iterated re-weighted least squares for the local optimization.
    \item Unlike the rest of the RANSAC implementations, kornia generates and evaluates hypotheses in "batches" of 1024 to make use of CPU and GPU parallelism.
\end{enumerate}\vspace{2mm}

\noindent \textbf{LO-RANSAC+ (pydegensac).} The algorithm from~\cite{fixingLORANSAC2012} as implemented in pydegensac 
package.It uses truncated quadratic cost function and fast local optimization scheme using a subset of inlier sets. 
\vspace{2mm}

\noindent \textbf{LO-RANSAC+ with LAF (pydegensac).} The additional component compared to the previous algorithm is the local affine frame check constraint proposed in~\cite{Mishkin2015MODS} for the fundamental matrix estimation.\vspace{2mm}

\noindent \textbf{GC-RANSAC (author).} The implementation provided by the authors. It uses a graph-cut-based local optimization that considers the spatial coherence of the 
input data points. 
The additional components are:
\begin{enumerate}[topsep=2mm, partopsep=1pt,itemsep=2pt,parsep=0pt]
    \item Sample cheirality check to reject minimal samples early if the implied plane flips between the two views.
    \item Single-sided re-projection error, measured in the second image, is used as point-to-model residual.
    \item Truncated quadratic cost function and fast iterative local optimization scheme. 
    \item Gaussian elimination for fast homography estimation from minimal samples.
    \item Column-pivoting QR decomposition for larger-than-minimal samples.
\end{enumerate}\vspace{2mm}

\noindent \textbf{GC-RANSAC with PROSAC (author).} The previous algorithm with PROSAC sampling~\cite{chum2005matching}.\vspace{2mm}

\noindent \textbf{GC-RANSAC (OpenCV) and MAGSAC++ (OpenCV).} The OpenCV implementation of the GC-RANSAC and MAGSAC++ algorithms.  Additional features:
\begin{enumerate}[topsep=2mm, partopsep=1pt,itemsep=2pt,parsep=0pt]
    \item Sequential Probability Ratio Test~\cite{chum2008optimal}.
    \item Gaussian elimination for fast homography estimation.
    \item Sample cheirality check to reject minimal samples early if the implied plane flips between the two views.
    \item Levenberg-Marquardt numerical optimization minimizes the re-projection error on the final set of inliers.
\end{enumerate}\vspace{2mm}

\noindent \textbf{RHO (OpenCV).} The OpenCV implementation of the method proposed in~\cite{bazargani2018fast}. The components are:
\begin{enumerate}[topsep=2mm, partopsep=1pt,itemsep=2pt,parsep=0pt]
    \item PROSAC sampling~\cite{chum2005matching}.
    \item Sequential Probability Ratio Test~\cite{chum2008optimal}.
    \item Gaussian elimination for fast homography estimation.
    \item Sample cheirality check to reject minimal samples early if the implied plane flips between the two views.
\end{enumerate}\vspace{2mm}

\noindent \textbf{MAGSAC (author) and MAGSAC++ (author).} The implementations provided by the authors. 
They use the following additional components for homography estimation. 
\begin{enumerate}[topsep=2mm, partopsep=1pt,itemsep=2pt,parsep=0pt]
    \item PROSAC sampling~\cite{chum2005matching}.
    \item Gaussian elimination for fast homography estimation from minimal samples.
    \item Column-pivoting QR decomposition for larger-than-minimal samples.
    \item Sample cheirality check to reject minimal samples early if the implied plane flips between the two views.
\end{enumerate}\vspace{2mm}

\noindent \textbf{VSAC (author).} The implementation provided by the authors.
They use the following additional components for homography estimation. 
\begin{enumerate}[topsep=2mm, partopsep=1pt,itemsep=2pt,parsep=0pt]
    \item Gaussian elimination for fast homography estimation from minimal samples.
    \item Householder QR decomposition for larger-than-minimal samples.
    \item Sample cheirality check to reject minimal samples early if the implied plane flips between the two views.
    \item Local optimization: non-minimal estimation on small subset of inliers (around 15-20 iterations)
    \item The MAGSAC++ optimization is applied in the end.  
\end{enumerate}\vspace{2mm}

\noindent \textbf{VSAC with PROSAC (author).} The previous algorithm using PROSAC sampling~\cite{chum2005matching}.\vspace{2mm}

\noindent \textbf{EAS (author).}
The implementation provided by the authors for the recently proposed algorithm in~\cite{fan2021efficient}. The method is implemented in pure Python using the numpy~\cite{numpy} package. 
\vspace{2mm}

\noindent \textbf{Affine-RANSAC (author).} The implementation provided by the authors for the method in~\cite{barath2020making} using affine correspondences to estimate the homography. 
Because our benchmark does not have affine correspondences, we approximate them using the SIFT features. 
Given rotations $\alpha_1, \alpha_2 \in [0, 2\pi]$ and scales $s_1, s_2$ in the two images for a correspondence, the affine transformation is calculated as $\textbf{A} = \textbf{J}_2 \textbf{J}_1^{-1}$, where $\textbf{J}_i = \textbf{R}_i \textbf{S}_i$, $\textbf{R}_i$ is the 2D rotation matrix rotating by $\alpha_i$ degrees, and $\textbf{S}_i$ is the 2D scale matrix uniformly scaling by $s_i$, $i \in [1, 2]$, along each axis.
They use the following additional components for homography estimation. 
\begin{enumerate}[topsep=2mm, partopsep=1pt,itemsep=2pt,parsep=0pt]
    \item SVD decomposition for estimating the homography affine correspondences.
    \item Sample cheirality check adapted for affine correspondences to reject minimal samples early.
    \item Graph-Cut RANSAC is used as robust estimator exploiting affine correspondences. 
\end{enumerate}\vspace{2mm}

\subsection{Deep Pre-filtering}

The standard two-view matching pipeline with SIFT or other local features uses SNN ratio test~\cite{SIFT2004} to filter-out unreliable correspondences before running RANSAC, otherwise, the inlier ratio is too small to have good results~\cite{IMC2020, brachmann2019ngransac, Efe_2021_ICCV}. Recently, it was shown~\cite{cne2018, dfe2018, brachmann2019ngransac, oanet2019, acne2020, clnet2021} that using neural networks for correspondence pre-filtering might provide significant benefits over the SNN ratio.

We evaluated how using such models for correspondence pre-filtering for uncalibrated epipolar geometry help homography estimation algorithms. 
For our study, we took pre-trained models, provided by the authors of each paper and use them for scoring the correspondences. 
We emphasize that we have neither trained, nor fine-tuned them for the homography estimation task, so their performance is sub-optimal compared to the same models, but trained for the homography estimation. 
The reason why we did not take the pre-trained models for homography is that authors do not provide them.
Since the sought homographies represent 3D planes in the COLMAP reconstruction, they stem from static structures.
The homography is thus consistent with the epipolar geometry of the static background.
Thus, filtering the correspondences with deep networks trained on epipolar geometry estimation reduces the outlier ratio also for homographies and makes the robust estimation easier.
%
Unless stated otherwise, all the pre-trained models we used, were trained on subset~\cite{cne2018} of YCC100M dataset correspondences for fundamental matrix estimation.\vspace{2mm}


\noindent \textbf{CNe}~\cite{cne2018}.
Context normalization networks (CNe) is the first paper on the topic which proposed to use PointNet (MLP) with batch normalization~\cite{batchnorm2015} as "context" mechanism. The model does not use any side information and the input is just a set of pair of coordinates in both images.\vspace{2mm}

\noindent \textbf{ACNe}~\cite{acne2020}.
Attentive context normalization networks introduces a special architectural block for the task. The model does not use any side information.\vspace{2mm}

\noindent \textbf{DFE}~\cite{dfe2018}.
Deep Fundamental matrix estimation uses differentiable iterative re-weighted least squares for the epipolar geometry estimation and the model predicts weights. It uses the following side information in addition to the point coordinates: difference in scale and orientation of the SIFT features, SNN ratio score, absolute descriptor difference score. 
Different from the rest of models, DFE was trained on Tanks and Temples dataset, which is smaller and less diverse in terms of camera poses than YCC100M dataset.\vspace{2mm} 

\noindent \textbf{OANet}~\cite{oanet2019}.
The OANet algorithm introduced several architectural blocks for the correspondence filtering estimation. It also uses the SNN ratio value and mutual nearest neighbor check as a side information.\vspace{2mm}

\noindent \textbf{Neural guiding}~\cite{brachmann2019ngransac}.
Neural-guided RANSAC paper uses a CNe-like architecture, but different training objective (reinforcement learning) and way of utilizing correspondence scores -- to perform importance sampling in the RANSAC. Note that we do not use the full NG-RANSAC as proposed in the paper, because there is no author implementation of it -- only the fundamental and essential matrix estimation. Instead, we only use the pre-trained model that scores the correspondences. It uses SNN ratio as a side information.\vspace{2mm}

\noindent \textbf{CLNet}~\cite{clnet2021}. 
CLNet introduces algorithmic and architectural advancement to first remove gross outliers but iterative pruning and only then look for the inlier candidates. No side information is used.

\section{Uncertainty of Keypoints}

The evaluation aims to determine bias and variance of angular, scale, and positional transformations of detected correspondences of SIFT keypoints $\{C_i\}_{i=1}^N$. Such statistics calculated on the same dataset allow comparison of different implementations of SIFT detectors. Moreover, we can compare the uncertainty of keypoints orientation, scale, and positions for any detector if such measurements are provided. We followed~\cite{barath2017theory} to derive an affine transformation (4 DoF) $\widetilde{\mat A}_i \in \mathbb{R}^{2 \times 2}$ in the vicinity of the keypoint pair from the reference homography $\widetilde{\mat H}_i$. The following sections discuss the evaluation of the positional differences, the determination of the reference scale ratios $\widetilde{r}_i$ and of the reference rotations $\widetilde{\alpha}_i$ and the transformation errors in detail. All the evaluations are measured on the OpenCV implementation of the SIFT detector.

\subsection{The positional transformation uncertainty}

The symmetric positional residual of each keypoint pair depends on the mean reprojection error
\begin{equation} \label{eq:epsilon-i}
\epsilon_{x_i} = \sqrt{(
|\mat x'_i-{\widetilde{\cal H}}(\mat x_i)|_2^2+|\mat x_i-\widetilde{{\cal H}}^{-1}(\mat x_i')|_2^2)/8} 
    .
\end{equation}

\begin{figure}
    \centering
    \includegraphics[width=0.5\textwidth]{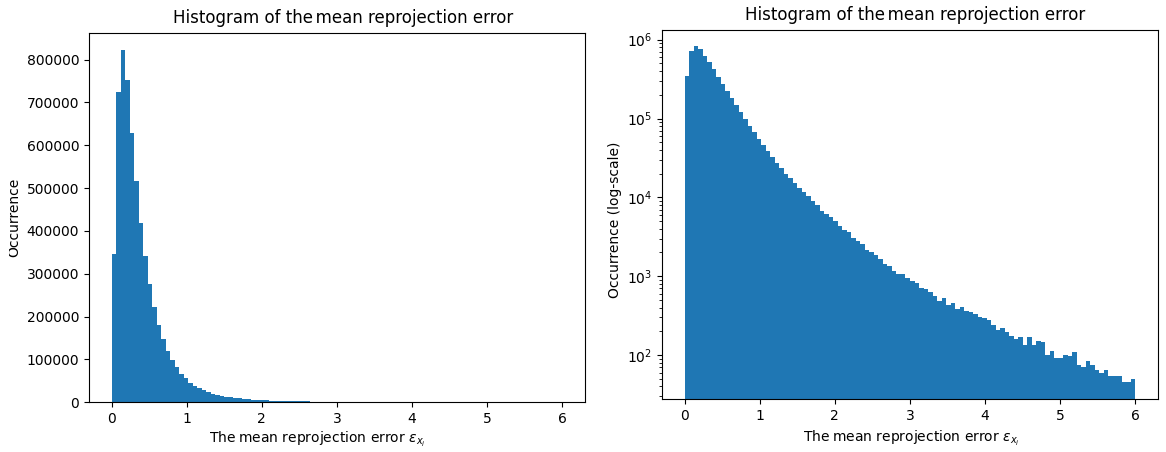}
    \caption{The residuals $\epsilon_{x_i}$ of $6.1$M keypoint pairs. The right histogram shows the logarithmic scale of the occurrence to visualize the distribution of the residuals. Measured standard deviation $\hat{\sigma}_{x} \approx 0.67$ pixels. The STD is a factor two larger, than expected, which might result from accepting small outliers.}
    \label{fig:his-positional-errors}
\end{figure}

\begin{figure}
    \centering
    \includegraphics[width=0.3
    \textwidth]{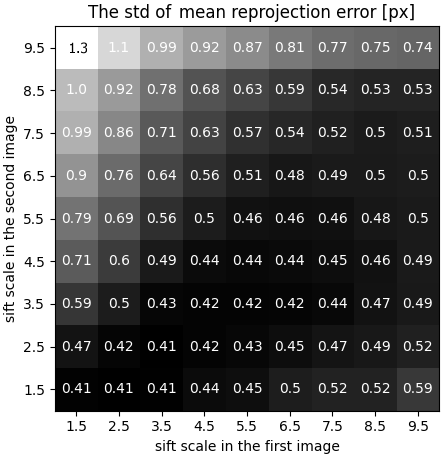}
    \caption{The standard deviation of $\epsilon_{x_i}$ for individual scale $s_i$,$s_i'$ combinations. We can see the dependence of reprojection accuracy on the scale of the related keypoints. }
    \label{fig:std-positional-errors-clustered}
\end{figure}

The histogram of residuals $\epsilon_{x_i}$ of $6.1$M keypoint pairs is in Figure~\ref{fig:his-positional-errors}. 
Furthermore, the authors W. Förstner and B. P. Worbel~\cite{forstner2016photogrammetric} show that the standard deviation of the keypoint depends on the detector scale (see \cite{forstner2016photogrammetric} p.681, \cite{zeisl*09:estimation} Eq.(15)). Therefore, it is reasonable to assume that the positional transformation error $\epsilon_{x_i}$ also depends on keypoint scales $s_i$, $s_i'$. We clustered the symmetric positional residuals w.r.t.~related $s_i$, $s_i'$ scales and measured the standard deviation for individual bins, see Figure~\ref{fig:std-positional-errors-clustered}.

\subsection{The scale transformation uncertainty}

\begin{figure}
    \centering
    \includegraphics[width=0.5\textwidth]{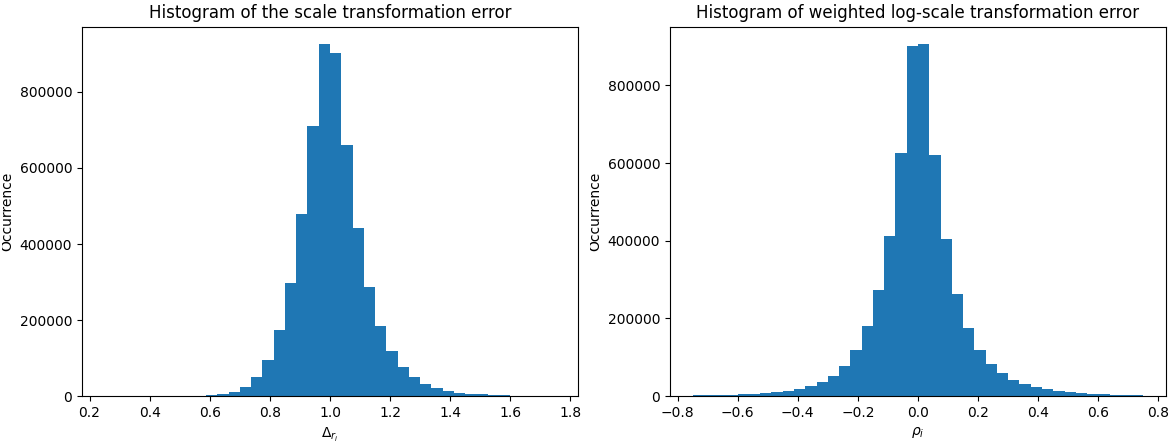}
    \caption{The histogram of the scale transformation ratio $\Delta r_i$ and the weighted log-ratio $\rho_i$ on 5.6M keypoint pairs.}
    \label{fig:his-scale-transform}
\end{figure}

The scale transformation uncertainty is evaluated using the ratios $r_i=s_i'/s_i$ (not to be confused with the redundancy numbers in the main paper) with the scales $(s_i, s_i')$ from the SIFT keypoints. The scale transformation accuracy is based on the ratio $\Delta r:= r_i/\tilde r_i$, where the ground truth scale ratio is derived from $\tilde {\mat A}$ via 
\begin{equation}
    \widetilde{r}_i = \sqrt{|\widetilde{\mat A_i}|} .
\end{equation}
%
%
For the cases with the affinity matrix having a condition number $ > 1.5$, the shears are assumed to have a too large impact on the scales. We only analyze cases with small scale ratios, \ie, assume values $\tilde r_i \in [0.5, 2]$. This interval contains $99.62\%$ keypoint pairs. Further, the weighted log-ratio $\rho_i = \log(\Delta r_i) / {\widetilde{r}_i}$ is calculated using the filtered $\Delta  r_i$ related to the ground truth $\widetilde{r}_i$. The scale statistics of the remaining 5.6M keypoint pairs are shown in Figure~\ref{fig:his-scale-transform}. 

\subsection{The angular transformation uncertainty}
\begin{figure}
    \centering
    \includegraphics[width=0.5\textwidth]{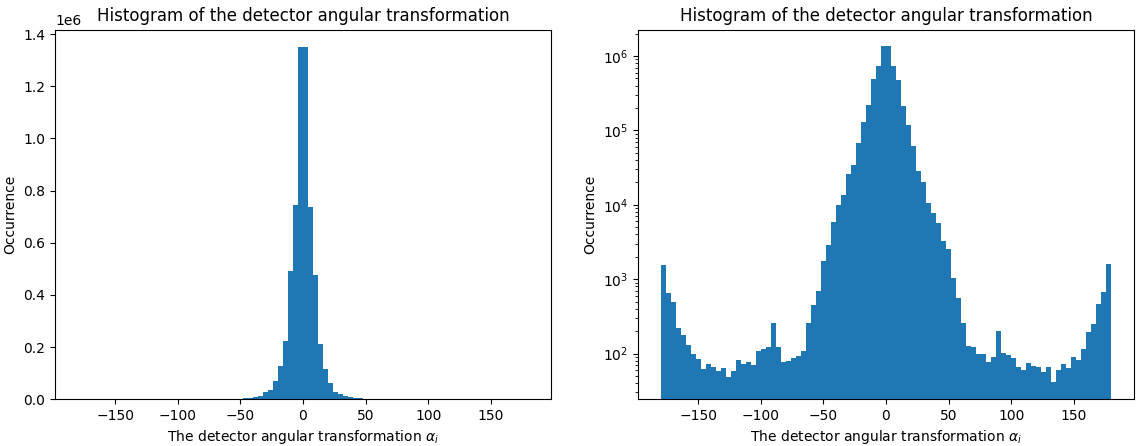}
    \caption{The histogram of the detector angular transformation $\alpha_i$ for 6.1M of keypoint pairs. The right histogram shows logarithmic scale of the occurrence to visualize the number of samples across the complete interval $[-180, 180)$ degrees.}
    \label{fig:his-angular-transform}
\end{figure}

The histogram of angular transformation $\alpha_i=\phi'_i-\phi_i$ for all keypoint pairs is visualized in Figure~\ref{fig:his-angular-transform}. The uncertainty of this transformation can be calculated by: (1) comparing direction vectors $\mat d(\phi'_{i})$ with the transformed direction $\mat d(\phi_{i})$ into the coordinates of $\mat d(\phi'_{i})$ or (2) deriving a local rotation from the reference homography and comparing it to the keypoint angular transformation.

\subsubsection{Comparing direction vectors}
\begin{figure}
    \centering
    \includegraphics[width=0.5\textwidth]{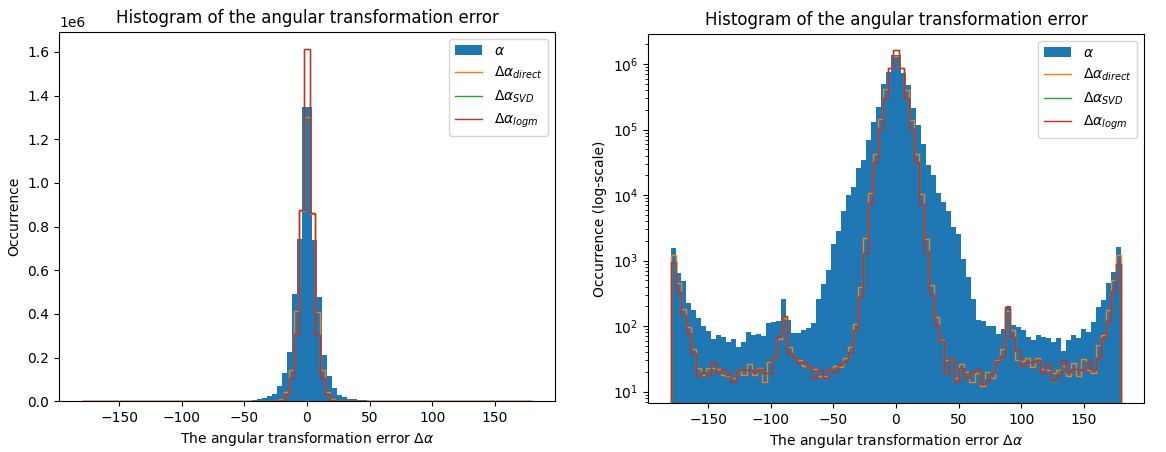}
    \caption{The histogram of angular transformation error $\Delta \alpha$ on top of $\alpha$. The transformation was (1.) measured as the angle between directional vectors, $\Delta \alpha_{{\scriptsize \mbox{direct}}}$ eq. (\ref{eq:alpha_direct})
    , (2.) subtracting the reference angular transformation decomposed by SVD, $\Delta \alpha_{{\scriptsize \mbox{SVD}}}$ eq. (\ref{eq:R_SVD}), and (3.) subtracting the ground truth angular transformation obtained from the exponential analysis, $\Delta \alpha_{{\scriptsize \mbox{logm}}}$ eq. (\ref{eq:rotational_component_exp}). We assumed $4.3$M correspondences with $\mbox{cond}(\widetilde{\mat A}_i) < 1.2$. The standard deviation is $\hat{\sigma}_{\alpha} \approx 7.9^{\circ}$ is approximately two times the rounding error.}
    \label{fig:his-angular-transform-err}
\end{figure}

The directional vector $\mat d_i = [\cos(\phi_i) \; \sin(\phi_i)]^\text{T}]$ realizing the first keypoint orientation can be transformed into the second image by the multiplication with the local approximation of affinity transformation (4DoF)
\begin{equation}
    \bar{\mat d}_i = \widetilde{\mat A}_i \mat d_i .
\end{equation}
The multiplication with the local affinity $\widetilde{\mat A}_i \in \mathbb{R}^{2 \times 2}$ does not include the projective part. The angle in the interval $[-\pi, \pi]$ can be obtained by
\begin{equation} \label{eq:alpha_direct}
    \Delta \alpha_{{\scriptsize \mbox{direct}}_i} = 
    \angle (\mat d_i', \bar{\mat d}_i) =
    \mbox{atan2}( | [\mat d_i', \bar{\mat d}_i] | ,  \mat d_i'^T \bar{\mat d}_i ) .
\end{equation}
This is a reasonable measure for evaluating the quality of the directions since -- assuming no outliers -- the expected value of this angular difference is zero,\footnote{Stochastical variables are underscored}  $\mbox{E}(\underline{\Delta \alpha}_{{\scriptsize \mbox{direct}}_i}) =0$.
Fig.~\ref{fig:his-angular-transform-err} shows the histogram of the $\Delta \alpha_{{\scriptsize \mbox{direct}}_i}$ from (\ref{eq:alpha_direct}).

\subsubsection{Partitioning of an affinity}

We assume $\widetilde{\mat A}_i \in \mathbb{R}^{2 \times 2}$ matrix locally approximate the homography $\widetilde{\mat H}_i \in \mathbb{R}^{3 \times 3}$. The goal of comparing SIFT directions could be to determine the rotation component $\tilde{\mat R}$ of the affinity $\widetilde{\mat A}_i$ and compare it to the angle between the directions of corresponding keypoints.

We address three alternatives for determining the rotational component of $\tilde{\mat A}$:
\begin{enumerate}
    \item a QR-decomposition,
    \item a SVD-decomposition, and 
    \item an exponential decomposition.
\end{enumerate}

\paragraph{Rotation from QR-decomposition of an affinity $\mat A$.}

Assuming the affinity is a concatenation of a shear matrix $\mat S$ and a subsequent rotation with $\mat R$
\begin{equation}
    \mat A = \mat R \mat S
\end{equation}
the classical QR-decomposition is defined as
\begin{equation}
    \mat R_{\mbox{\tiny qr,A}}:= \mat R \quad \mbox{with} \quad [\mat R, \mat S] = \qr(\mat A)\,.
\end{equation}
In case the affinity is defined by the reverse sequence, \ie
\begin{equation}
    \mat A= \mat S \mat R
\end{equation}
the QR decomposition of the transposed needs to be taken
\begin{equation}
    \mat R_{\mbox{\scriptsize qr,A$\trans$}}:= \mat R\trans\quad \mbox{with} \quad [\mat R, \mat S] = \qr(\mat A\trans)\,.
\end{equation}
If there are no shears, \ie the shear matrix is a scaled unit matrix, the two rotations $\mat R_{\mbox{\scriptsize qr,A}}$ and
$\mat R_{\mbox{\scriptsize qr,A$\trans$}}$ are the same, otherwise they differ.

\paragraph{Rotation from SVD-decomposition of $\mat A$.}

An alternative way to derive the rotation component uses the matrix exponential. Let us assume, the affinity is decomposable as two rotations sandwiching a individual scaling 
\begin{equation}
    \mat A = \mat U \mat {D} \mat V\trans \quad \mbox{with} \quad \mat D = \mx{cc}{d_1 &  0 \\0&  d_2}\,,
\end{equation}
where the shears are represented by the rotation $\mat V$ and the ratio $d_1/d_2$.
Then the SVD yields the rotation
\begin{equation} \label{eq:R_SVD}
    \mat R_{\mbox{\scriptsize svd,A}}:= \mat U \mat V\trans \quad \mbox{with} \quad [\mat U,  \mat {\Lambda}, \mat V] = \svd(\mat A)\,.
\end{equation}
Transposing $\mat A$ does not change the rotation.
The resulting rotation only is identical to those of the QR-decomposition if the affinity is a scaled rotation.

\paragraph{Rotation from an exponential decomposition}

The affinity $\mat A$ can be written as an exponential of a matrix $\mat B$
\begin{equation}
    \mat A = \mbox{e}^{\mat B}
\end{equation}
If the matrix $\mat B$ is zero, i.e. $\mat B= \mat 0$, the affinity is a unit transformation. We now can decompose the exponent additively in the following form
\begin{equation}
    \mat B = \sum_i p_i \mat B_i
\end{equation}
with the four basic $2\times 2$ matrices
\begin{eqnarray}
    \mat B_1  =&\mx{cc}{1 & 0\\0 & 1} \,,\quad \mat B_2 &= \mx{cc}{0 & -1\\1 & 0} \\
    \mat B_3 &= \mx{cc}{0 & 1\\1 & 0} \,,\quad \mat B_4 =&\mx{cc}{1 & 0\\0 & -1}  \,.
\end{eqnarray}
Hence
\begin{equation}
    \mat A = \mbox{e}^{p_1 \mat B_1 + p_2 \mat B_2 + p_3 \mat B_3 + p_4 \mat B_4 }\,.
\end{equation}
If we take each of the summands individually, 
the four parameters refer to (1) scaling with $\log p_1$, (2) rotation by $p_2$ [rad], (4) 1st shear, namely opposite scaling of axes, and (4) 2nd shear, namely opposite rotation of axes. The rotation is given by the well known relation 
\begin{equation}
    \label{eq:rotational_component_exp}
    \mat R=\exp(p_2\mat B_2)\,.
\end{equation}
Furthermore, for the first shear we explicitely have
\begin{eqnarray}
    &&\exp\left( \mx{cc}{0 & p_4\\ p_4 & 0}\right) 
    \\&=&
    \mx{cc}{ 
    \mbox{e}^{-p_4/2} + \mbox{e}^{p_4/2} & \mbox{e}^{p_4/2} - \mbox{e}^{-p_4/2}\\
    \mbox{e}^{-p_4/2} - \mbox{e}^{-p_4/2} & \mbox{e}^{-p_4/2} + \mbox{e}^{p_4/2}
} \\
&\stackrel {q_4 = \mbox{\scriptsize e}^{p_4/2}}=&
    \mx{cc}
    {q_4+1/q_4 & q_4-1/q_4 \\ q_4-1/q_4 & q_4+1/q_4
    }\,.
\end{eqnarray}
This representation is highly symmetric. The additive terms are invariant w.r.t. the sequence of the terms. 
Moreover, the scaled rotation is independent on the existence of shears.

However, since the exponent of two matrices only is the product of the two matrices if they commute, i.e.
\begin{equation}
    \exp(\mat A+\mat B)=\exp(\mat A) \exp(\mat B) \quad \mbox{only if} \quad \mat A\mat B = \mat B \mat A\,,
\end{equation}
the interpretation of the elements in the exponent is not independent of the existence of the other elements.
Only a common scaling can be exchanged with the other components, as is known from scaled rotation.

Now, we can define the rotational component using (\ref{eq:rotational_component_exp}) deriving $p_2$ from
\begin{equation}
    p_2 = (B(2,1)-B(1,2))/2  \quad \mbox{with} \quad \mat B = \log(\mat A)
\end{equation}
where $\log(\mat A)$ is the matrix logarithm of $\mat A$.

Therefore we are able to identify the existence of shears, namely we have no shears if
\begin{equation}
    d_s^2= |[p_3,p_4]|=p_3^2+p_4^2 = 0
\end{equation}
Since a scale rotation has condition number $\mbox{cond}(s\mat R)=1$, also the condition number can be used to identify the lack of shears, namely if \mbox{cond}$(\mat A)=1$. For not too large shears the the condition number and the  degree of shears $d_s^2$ are approximately the same:
\begin{eqnarray}
    d_s^2 \approx \mbox{cond}(\mat A)\,.
\end{eqnarray}

\section{Effect of Weighting and Estimation Type}

\subsection{Outline of the analysis}

We use two sets of sample data to answer two questions:
\begin{enumerate}
    \item What loss in accuracy is to be expected when using an algebraic estimation vs. a ML-estimation?
    \item What effect on the accuracy does a scale dependent weighting have onto the results of an ML-estimation? (see \cite{forstner2016photogrammetric}, Sect. 15.4.1.3)
\end{enumerate}
The first set A was chosen, such that (1) the number of correspondences is small in order to allow for non-uniform distribution of points and (2) the shears to be large, the planes are not fronto-parallel in order to have the homographies largely deviate from a scaled rotation. The second set B is the same as been used for the investigation into the uncertainty of the SIFT detector.

\subsection{Algebraic and ML minimization}
We apply two estimation methods, each yielding covariance matrices for the homography parameters based on the constraints using the observations, containing the homogeneous coordinates of the keypoint pairs
$\mat l_i:=[\mat x_i, \mat x_i']^T$ with the covariance matrix $\mat \Sigma_{l_il_i}$ and 
the unknown parameters $\vunn  := \mbox{vec} \mat H $ and $\mat y_i =\mathbb{E}(\mat l_i)$
\begin{equation}
    \mat 0 = \mat g_i(\vunn,\mat y_i)):= \mathbb{E}(\mat x_i'^T)  \times (\mat H \;\mathbb{E}(\mat x_i))
\end{equation}
which linearized has the form
\begin{equation}\label{eq:g-linear}
    \mat g_i(\vunn,\mat y_i)=\mat g_i(\vunn^0,\mat y^0_i) + \mat A_i \Delta \vunn + \mat B_i^T \Delta \mat y_i = \mat 0\,.
\end{equation}
with the Jacobians
\begin{equation}
    \mat A_i = \frac{\partial \mat g_i}{\partial \vunn} \quad \mbox{and} \quad \mat B_i^T = \frac{\partial \mat g_i}{\partial \mat l_i}\, ,
\end{equation}
yielding the complete observation vector, constraints, and  complete Jacobians 
\begin{equation} \nonumber
    \mat l=[\mat l_i] \,, \quad  \mat g=[\mat g_i]    \,, \quad \mat A = [\mat A_i] \,,\quad \mbox{and} \quad \mat B = \mbox{Diag}(\mat B_i^T)\, ,
\end{equation}
i.e.  using the block diagonal matrix Diag($\cdot$) with the $\mat B^T_i$ as entries. We obtain the following covariance matrices for the homography parameters.
\begin{enumerate}
    \item The {\em classical algebraic method} minimizing the algebraic error
    \begin{equation}
        \Omega^{\mbox{\tiny (ALG)}}(\vunn) = \mat g^T(\vunn,\mat l) \mat g(\vunn, \mat l) \,,
    \end{equation}
    yields the linear relation from (\ref{eq:g-linear})
    \begin{equation}\label{eq:x-alg-l}
        \widehat{\Delta \vunn} =- (\mat A^T\mat A)^{+} \mat A^T  \mat B^T \Delta \mat l\,,
    \end{equation}
    see \cite{forstner2016photogrammetric}, eq. (4.518), from which we obtain the covariance matrix
    \begin{equation}
        \mat \Sigma_{\hat {\unn}\hat {\unn}}^{\mbox{\tiny (ALG)}} =  (\mat A^T\mat A)^{+} \mat A^T \mat B^T  \mat \Sigma_{ll} \mat B\mat A  (\mat A^T\mat A)^{+} \,.
    \end{equation}
    Observe, we only would obtain the covariance matrix $(\mat A^T \mat A)^{-1}$ if the covariance matrix $\mat B^T  \mat \Sigma_{ll} \mat B$ of the constraints $\mat A \Delta \vunn$ would be the unit matrix, see (\ref{eq:g-linear}), which generally does not hold.
    
    \item The {\em ML-estimation}, taking the uncertainty of the points into account, minimizes
    \begin{equation}
        \Omega^{\mbox{\tiny (ML)}}(\vunn) = \mat v^T\mat \Sigma_{ll}^{-1}\mat v\,
    \end{equation}
    with $\mat v = \mat y-\mat l= (\mat y^0+\Delta \mat y)-\mat l$,
    under the constraints (\ref{eq:g-linear}), which include the unknown parameters $\vunn$,
    and yields the linear relation  
    \begin{equation}\label{eq:x-ml-l}
        \widehat{\Delta \vunn} = -\mat \Sigma_{\hat {\unn}\hat {\unn}}^{\mbox{\tiny (ML)}}  \mat A^T (\mat B^T \mat \Sigma_{ll} \mat B)^{-1}  \mat B^T\Delta \mat l\,,
    \end{equation}
    see \cite{forstner2016photogrammetric}, eq. (4.447) with the covariance matrix
    \begin{equation}
        \mat \Sigma_{\hat {\unn}\hat {\unn}}^{\mbox{\tiny (ML)}} = (\mat A^T (\mat B^T \mat \Sigma_{ll} \mat B)^{-1} \mat A)^{+}\,.
    \end{equation}
    If we assume $\mat B^T \mat \Sigma_{ll} \mat B=\mat I$, which is what algebraic minimization does, eq. (\ref{eq:x-ml-l}) reduces to obtain (\ref{eq:x-alg-l}).
    
\end{enumerate}
In both cases we do not make the procedural details explicit, which are caused by the redundant representation of the homography and the homogeneous coordinates: Actually, the covariance matrix $\mat \Sigma_{\hat {\unn}\hat {\unn}}$ has rank 8, since the homography only has 8 d.o.f., similarly, the covariance matrix of a homogeneous vector $\mat x$ representing a 2D point, is rank 2. In both cases, we employ a minimal representation in the tangent space defined by the constraints $||\mat H||_2=1$ and $|\mat x|_2=1$. Details for an ML-estimation of a homography are given in \cite{forstner2016photogrammetric}, Sect. 10.6.3.

\subsection{Scale dependent weighting}
We use two different weighting schemes for the ML-estimation
\begin{enumerate}
    \item Equal weights for all points
    \begin{equation} \label{eq:w=1}
        w_{1}(i)= w_0(i')=1\,.
    \end{equation}
    \item Choosing the weights as a function of the scales of the $I$ keypoints, namely
    \begin{equation}
        w_{s}(i)= \frac{m^2}{s^2(i)} \quad \mbox{and} \quad
        w_{s}(i')= \frac{m^2} {s^2(i')}\,,
    \end{equation}
    with the geometric mean of all scales 
    \begin{equation} 
        m = \left(\prod_i s_i \prod_{i'} s_{i'}\right)^{1/(2I)}\,.
    \end{equation}
    The denominator is meant to have the average variance 1, to be comparable to (\ref{eq:w=1}), though the results do not depend on this common scaling.
    
    For the ML-estimates, in addition to the covariance matrices $\mat {\Sigma}_{\hat {\unn}\hat {\unn}}$ we also obtain the estimated variance factor
    \begin{equation}
        \hat{\sigma}_0^2 = \frac{\Omega(\hat {\vunn})}{R}
    \end{equation}
    which depends on the weighted sum $\Omega$ of the squared residuals, i.e. the reprojection errors and the redundancy $R=2I-8$ of the estimation. It tells by which factor we need to multiply the assumed covariance matrix in order to obtain an unbiased covariance matrix, assuming the given covariance matrix provides the correct ratio of the uncertainties between the observations:
    \begin{equation}
        \mat \Sigma_{\hat {\unn}\hat {\unn}}^{\mbox{\tiny a posteriori}} = \sigma_0^2 \mat \Sigma_{\hat {\unn}\hat {\unn}} ^{\mbox{\tiny a priori}}\,.
    \end{equation}
\end{enumerate}

\subsection{Accuracy Evaluation criteria}
We use the following {\em criteria} to determine the loss in accuracy, i.e. an increase of the standard deviations $\sigma_{\hat {\unn}_u}$, when comparing the covariance matrix $\mat \Sigma$ to a reference covariance matrix $\widetilde{\mat \Sigma}$,  namely the mean loss
\begin{equation}
    l_{\mbox{\tiny mean}} = \sqrt{   \text{trace}(\mat \Sigma_{\hat {\unn}\hat {\unn}}  \widetilde{\mat \Sigma}_{\hat {\unn}\hat {\unn}}^{-1})/8 }
\end{equation}
and the maximum loss
\begin{equation}
    l_{\mbox{\tiny max}} = \sqrt{ \max \lambda(\mat \Sigma_{\hat {\unn}\hat {\unn}} \widetilde{\mat \Sigma}_{\hat {\unn}\hat {\unn}}^{-1}) }\,,
\end{equation}
see \cite{foerstner*17:efficient}. In case the two matrices are diagonal matrices with the variances, we obtain {\em the average and the maximum ratio of the standard deviations}. 

\section{Geometry and Statistics for Sect. 5}

\subsection{On the estimate $\epsilon_{x_i}$ for $\sigma_i$}

We show, that 
\begin{equation}\label{eq:epsilon}
    \epsilon_{x_i} = \sqrt{(|\mat x_i - {\cal H}_i (\mat x_i')|_2^2 + |\mat x_i' - {\cal H}_i^{-1} (\mat x_i)|_2^2)/8}
\end{equation}
is a meaningful estimate for the standard deviation $\sigma_i$ of all coordinates $u_{ij}$ and $u'_{ij}$ of the given points
$\mat x_i =(u_{i1}, u_{i2})$ and $\mat x_i' =(u'_{i1}, u'_{i2})$.
Hence, we assume$\mbox{D}(\smat e_i)= \mbox{E}(\smat e_i\smat e_i\trans)=\mbox{D}(\smat e_i')=\mbox{E}(\smat e_i'\smat e\transs_i) = \sigma_i^2 \mat I_2$, which holds for the errors $\smat e_i=\smat u_i-\mbox{E}(\s{\mat u}_i)$ and $\smat e_i'=\smat u_i'-\mbox{E}(\smat u_i')$. Linearizing $\mat x_i - {\cal H}_i (\mat x_i')$ leads to $\smat e_i - \mat A_i (\smat e_i')$, and similarly for the second term. Thus, the RMSE, \ie the expession under the squareroot  in (\ref{eq:epsilon}) is linearized to 
\begin{equation}
    \Omega_i =|\mat e_i - \mat A_i (\mat e_i')|_2^2 + |\mat e_i' - \mat A_i^{-1} (\mat e_i)|_2^2
\end{equation}
We now determine the expectation $\mbox{E}(\s{\Omega}_i)$ and obtain
\begin{eqnarray}
    \mbox{E}(\s{\Omega}_i)
    &=& \mbox{E}\left((\smat e_i - \mat A_i (\smat e_i'))\trans(\smat e_i - \mat A_i (\smat e_i')) \right.\\ 
    &&\left.+ (\smat e_i' - \mat A_i^{-1} (\mat e_i))\trans (\mat e_i' - \mat A_i^{-1} (\mat e_i))\right)\\
    &=& \mbox{E}\left(\smat e_i\trans \smat e_i + \smat e_i\transs \mat A_i\trans \mat A_i \smat e_i' \right.\\ 
    &&\left.+ \mat e_i\transs \smat e'_i + \smat e_i\trans \mat A_i^{-T} \mat A_i^{-1} \smat e_i\right)
\end{eqnarray}
With $\mbox{tr}(\mat U\mat V)=\mbox{tr}(\mat V\mat U)$, thus $\mat a\trans\mat S\mat a=\mbox{tr}(\mat a\trans\mat S\mat a)=\mbox{tr}(\mat S\mat a\mat a\trans)$ we then obtain
\begin{eqnarray}
    \mbox{E}(\s{\Omega}_i)
    &=& \mbox{E}\left(\smat e_i\trans \smat e_i + \smat e_i\transs \mat A_i\trans \mat A_i \smat e_i' \right.\\ 
    &&\left.+ \smat e_i\transs \smat e'_i + \smat e_i\trans \mat A_i^{-T} \mat A_i^{-1} \smat e_i\right)\\ 
    &=& \mbox{E}\left(\mbox{tr}(\smat e_i\smat e_i\trans) + \mbox{tr}(\mat A_i\trans \mat A_i \smat e_i'\smat e_i\transs ) \right.\\ 
    &&\left.+ \mbox{tr}(\mat e_i'\mat e_i\transs) + \mbox{tr}( \mat A_i^{-T} \mat A_i^{-1} \smat e_i\smat e_i\trans\right)\\ 
    &=& \mbox{tr}(\mbox{E}(\smat e_i \smat e_i\trans)) + \mbox{tr}(\mat A_i\trans \mat A_i \mbox{E}(\smat e_i'\smat e_i\transs) ) \\ 
    &&+\mbox {tr}(\mbox{E}(\smat e_i' \smat e_i\transs)) + \mbox{tr}( \mat A_i^{-T} \mat A_i^{-1} \mbox{E}(\mat e_i\mat e_i\trans))\\ 
    &=& \mbox{tr}(\mat I_2) \sigma_i^2 + \mbox{tr}(\mat A_i\trans \mat A_i)\sigma_i^2 \\ 
    &&+\mbox{tr}(\mat I_2) \sigma_i^2 + \mbox{tr}( \mat A_i\transi\mat A_i^{-1} )\sigma_i^2\\ 
    &=& (4+\mbox{tr}(\mat A_i\trans \mat A_i)+\mbox{tr}( \mat A_i\transi \mat A_i^{-1} )) \sigma_i^2 \,.   
\end{eqnarray}
With the eigenvalues $\lambda_{1,2}(\mat A_i\trans \mat A_i)$ we now obtain    
\begin{equation}
    \mbox{E}(\s{\Omega}_i)   = (4+\lambda_1 + \lambda_2+ 1/\lambda_1 +1/\lambda_2) \sigma_i^2 \ge 8 \sigma_i^2
\end{equation}
since $x+1/x = (1-x)^2/x + 2 \ge 2$ for $x>0$. Hence, if $\lambda_1=\lambda_2=1$, thus for a pure rotation, the value $\epsilon_{x_i}^2$ is an unbiased estimator for $\sigma_i^2$. Dividing the RMSE $\Omega$ by $\sqrt 8$ therefore leads to a conservative estimate of the standard deviation $\sigma_i$.

\subsection{Affinity and Slope of Plane}

We give a relation between the condition number and the slope of a plane observed by an image pair in normal position.

The image of a sloped plane leads to scale differences $s$ and shears $a$ due to the tilts $Z_x$ and $Z_y$ of the plane along and across the base line. They have the form
 \begin{equation}
     \mat A_s = \left[\begin{array}{cc}
         1+s &  0\\
        0  & 1
     \end{array}
     \right]
     \quad \mbox{and} \quad
     \mat A_a = \left[\begin{array}{cc}
         1 &  a\\
        0  & 1
     \end{array}
     \right]
 \end{equation}
 The combined effect is the affinity
 \begin{equation}
     \mat A_{sa} = \mat A_{a} \mat A_{s}= \left[\begin{array}{cc}
         1+s &  a\\
        0  & 1
     \end{array}
     \right]
 \end{equation}

 \paragraph{Condition Number for Affinity except Scaled Rotation.}
 The condition number of this affinity, is given by
\begin{equation}
    c = 1+ \frac{(\sqrt{4(1+s)+t^2}+t)}{2(1+s)} t \quad \mbox{with} \quad t^2 = a^2+s^2
\end{equation}
For small $s$ and $t$ it can be approximated by
\begin{equation}
    c \approx 1 + t = 1 + \sqrt{a^2+s^2}
\end{equation}
neglecting higher order terms.

\paragraph{Affine Parameters and Slope of Scene Plane.}

Assume the stereo image pair in normal position with rotation $\mat R=\mat I$, basis $\mat b=[1,0,0]\trans$, and focal length $f=1$ with coordinate system in the first camera observing a sloped plane at $[0,0,Z_0]\trans$
\begin{equation}
    Z=Z_0+X Z_X + Y Z_Y \quad \mbox{with} \quad Z_X = \frac{\partial Z}{\partial X}\,, \; \; Z_Y = \frac{\partial Z}{\partial Y}
\end{equation}
or with homogeneous plane coordinates 
\begin{equation}
    \mat A = [Z_X,Z_Y,-1,Z_0]\trans = [\mat n\trans, Z_0]\trans\,.
\end{equation}
The homography from $\mat x'$ to $\mat x''$ is given by $\mat x''=\mat H\mat x'$ by
\begin{equation}
    \mat H =  \mat I + \frac{\mat b \mat n\trans}{Z_0} = 
    \left[\begin{array}{ccc} \frac{\mathrm{Z_X}}{Z_0}+1 & \frac{\mathrm{Z_Y}}{Z_0} & -\frac{1}{Z_0}\\ 0 & 1 & 0\\ 0 & 0 & 1 \end{array}\right]
\end{equation} 
This is an affinity with a Jacobian independent of the position in the image, namely
\begin{equation}
    \mat A = \frac{\partial \mat x''}{\partial \mat x'}=
    \displaystyle{\left(\begin{array}{cc} 1+\frac{Z_X}{Z_0} & \frac{Z_Y}{Z_0}\\ 0 & 1 \end{array}\right)}\,.
\end{equation}
Hence, we have the scale difference and the shear
\begin{equation}
    s= \frac{\mathrm{Z_X}}{Z_0} \quad \mbox{and} \quad a = \frac{\mathrm{Z_Y}}{Z_0}\,.
\end{equation}

\section{Results}

\subsection{Data set A}

\begin{figure}
    \centering
    \includegraphics[width=0.5\textwidth]{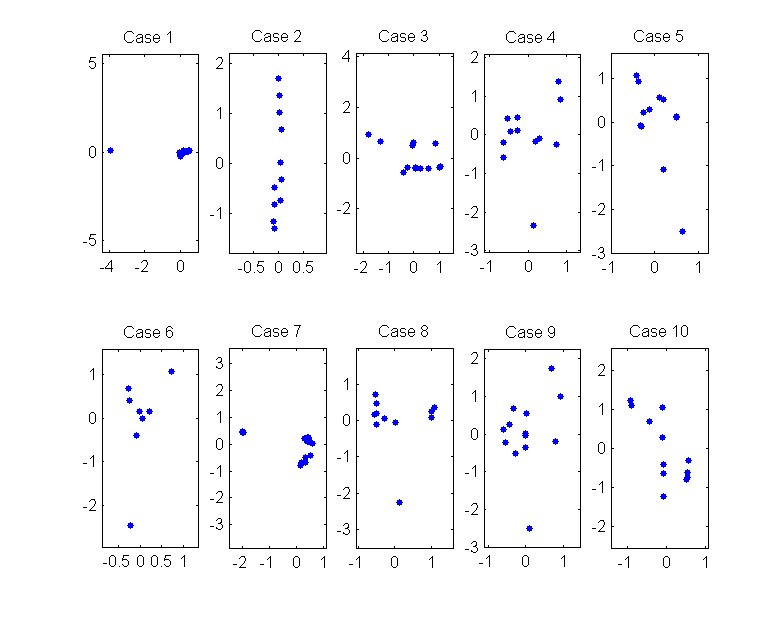}
    \caption{The distribution of the points for the 10 cases, see Tab. \ref{tab:results_10} }
    \label{fig:example_10_cases}
\end{figure}

For the 10 cases of data set A the results are collected in Table \ref{tab:results_10}.
Fig. \ref{fig:example_10_cases} shows the point distribution of the 10 cases. 
\begin{table*}
    \centering
    \begin{tabular}{c||r|rrr|rr|rr|rr}
    \hline
    case  &   $I$ &  $\min(\sigma_x)$ &   $\max(\sigma_x)$ &  $\displaystyle{\frac{\max(\sigma_x)}{\min(\sigma_x)}}$   & $\sigma_0(w=1)$ &  $\sigma_0(w(s))$ &  $l_{\mbox{\tiny mean}}^{\mbox{\tiny ALG$|$ML}}$   &  $l_{\mbox{\tiny max}}^{\mbox{\tiny ALG$|$ML}}$ &  $l_{\mbox{\tiny mean}}^{\mbox{\tiny 1$|$s}}$   & $l_{\mbox{\tiny max}}^{\mbox{\tiny 1$|$s}}$ \\
      & & [px] & [px] & &  [px] & [px] & & & \\ 
    \hline
     & 1 & 2 & 3 & 4 & 5 & 6 & 7 & 8 & 9 & 10  \\ 
    \hline\hline
 1 & 18 & 0.28 & 2.65 & 10 & {\bf 0.340} & {\bf 0.377} & 1.005 & 1.047 & 1.105 & 1.350\\
 2 & 12 & 0.32 & 2.52 &  8 & 0.170 & 0.179 & 1.052 & 1.318 & 1.223 & 1.569\\
 3 & 13 & {\bf 0.48} & 2.08 &  4 & 0.133 & 0.129 & 1.050 & 1.273 & 1.076 & 1.251\\
 4 & 14 & 0.29 & 4.66 & 16 & 0.266 & 0.172 & {\bf 1.105} & {\bf 1.418} & {\bf 2.343} & 4.098\\
 5 & 13 & 0.26 & {\bf 6.11} & {\bf 23} & 0.228 & 0.276 & 1.043 & 1.218 & 2.206 & {\bf 4.157}\\
 6 & 10 & 0.27 & 2.14 &  8 & 0.063 & 0.060 & 1.037 & 1.163 & 1.126 & 1.348\\
 7 & {\bf 24} & 0.26 & 2.59 & 10 & 0.252 & 0.287 & 1.002 & 1.009 & 1.131 & 1.280\\
 8 & 12 & 0.24 & 5.20 & 22 & 0.165 & 0.242 & 1.023 & 1.087 & 2.119 & 2.989\\
 9 & 16 & 0.29 & 5.62 & 20 & 0.261 & 0.285 & 1.102 & 1.331 & 1.976 & 3.382\\
10 & 14 & 0.24 & 5.36 & 22 & 0.203 & 0.296 & 1.054 & 1.264 & 1.860 & 2.972\\\hline\hline
mean &   &  0.29 &  3.89 & 14 &  0.208 &  0.230 &  1.047 &  1.213 &  1.616 &  2.440\\
max & 24 &  0.48 &  6.11 & 23 &  0.340 &  0.377 &  1.105 &  1.418 &  2.343 &  4.157
    \end{tabular}
    \caption{Data set A. Comparing the accuracy of different homography estimates. Columns: (1) Number $I$ of point pairs, (2--4) minimal and maximal standard deviations  $\min(\sigma_x)$ and $\max(\sigma_x)$ of weighted points and their ratio, (5--6) estimated variance factors for unweighted and weighted points, (7--8) mean and maximal losses when comparing the algebraic with the ML-estimate using equal weights, $l_{\mbox{\tiny mean}}^{\mbox{\tiny ALG$|$ML}}$ and  $l_{\mbox{\tiny max}}^{\mbox{\tiny ALG$|$ML}}$, and  (9--10) mean and maximal losses when comparing the unweighted and the weighted ML-estimate, $l_{\mbox{\tiny mean}}^{\mbox{\tiny 1$|$s}}$ and  $l_{\mbox{\tiny max}}^{\mbox{\tiny 1$|$s}}$, see Fig. \ref{fig:example_10_cases}}
    \label{tab:results_10}
\end{table*}

\paragraph{Discussion.}

The table allows the following conclusions:
\begin{itemize}
    
     \item The prior scale dependent standard deviations lies in a range between 0.24 pixel and 6.11 pixel. Since the redundancy in all cases is small, these values are quite uncertain. However, their ratios, which only depend on the scales of the points, are as uncertain as the scales are. The ratios vary between 4 in case 3, and 23 in case 5.
     
     \item Starting from an uncertainty of 1 pixel, the estimated (square rooted) variance factors $\hat{\sigma}_0$ indicate that on an average the keypoint coordinates are much better than 1 pixel, approximately by a factor 4 to 5. 
     
     \item The loss in accuracy when using the classical algebraic method for homography estimation compared to the achievable accuracy using a ML-estimation is shown in columns 8 and 9. The mean loss mostly is below 10\%, which appears acceptable. However, the maximum loss is about 42 \% (case 4). 
     
     \item The loss in accuracy when using equally weighted coordinates instead of taking the (assumed) scale dependency into account is shown in columns 10 and 11. While the mean loss lies between 8\% in case 3 and a factor 2.3 in case 4, the maximum loss reaches a factor 4.2 in case 5. The variation of the standard deviations (column 5) is approximately coherent with the loss in accuracy.
\end{itemize}

\subsection{Data set B}

Data set B consists of 969 homographies with alltogether 22~489 correspondences. We first provide the result of the first 30 cases in Tab. \ref{tab:experiment-large-1-30} together with the mean and maximum values for each criterion. The results confirm the findings of data set A, of course leading to more extreme ranges/maximum values.

\begin{table*}
    \centering
    \begin{tabular}{c||r|rrr|rr|rr|rr}
    \hline
    case  &   $I$ & $\min(\sigma_x)$ &   $\max(\sigma_x)$ &  $\displaystyle{\frac{\max(\sigma_x)}{\min(\sigma_x)}}$   & $\sigma_0(w=1)$ &  $\sigma_0(w(s))$ &  $l_{\mbox{\tiny mean}}^{\mbox{\tiny ALG$|$ML}}$   &  $l_{\mbox{\tiny max}}^{\mbox{\tiny ALG$|$ML}}$ &  $l_{\mbox{\tiny mean}}^{\mbox{\tiny 1$|$s}}$   & $l_{\mbox{\tiny max}}^{\mbox{\tiny 1$|$s}}$ \\
      & &  [px] & [px] & & [px] & [px] & & & \\ 
    \hline
     & 1 & 2 & 3 & 4 & 5 & 6 & 7 & 8 & 9 & 10   \\ 
    \hline\hline
1 & 39 &0.34 & 2.18 & 6 & 0.216 & 0.209 & 1.050 & 1.192 & 1.119 & 1.303\\
2 & 44 &0.31 & 6.96 & 23 & 0.341 & 0.358 & 1.042 & 1.131 & 2.511 & 4.451\\
3 & 34 & 0.37 & 5.07 & 14 & 0.337 & 0.302 & 1.146 & 1.619 & 1.688 & 2.267\\
4 & 33 & 0.17 & 1.82 & 10 & 0.238 & 0.287 & 1.080 & 1.510 & 1.054 & 1.111\\
5 & 17 & 0.16 & 4.83 & 31 & 0.395 & 0.460 & 1.038 & 1.090 & 2.112 & 3.086\\\hline
6 & 31 & 0.40 & 9.13 & 23 & 0.701 & 0.781 & {\bf 1.169} & {\bf 1.918} & 2.084 & 4.254\\
7 & 21 & 0.39 & 2.63 & 7 & 0.415 & 0.474 & 1.112 & 1.598 & 1.254 & 1.770\\
8 & 22 & 0.45 & 2.60 & 6 & 0.255 & 0.270 & 1.061 & 1.223 & 1.202 & 1.458\\
9 & 16 & 0.38 & 7.71 & 20 & 0.367 & 0.494 & 1.049 & 1.138 & 2.274 & 3.921\\
10 & 37 & 0.34 & 3.46 & 10 & 0.178 & 0.197 & 1.052 & 1.276 & 1.392 & 1.770\\\hline
11 & 14 & 0.33 & 3.54 & 11 & 0.097 & 0.117 & 1.045 & 1.308 & 1.062 & 1.214\\
12 & 30 & 0.51 & 1.91 & 4 & 0.158 & 0.161 & 1.061 & 1.224 & 1.087 & 1.209\\
13 & 36 & 0.24 & 8.76 & 37 & 0.334 & 0.325 & 1.007 & 1.030 & 3.255 & 5.093\\
14 & 26 & 0.36 & 4.14 & 12 & 0.302 & 0.296 & 1.016 & 1.052 & 1.828 & 2.487\\
15 & 42 & 0.19 & {\bf 14.79} & {\bf 77} & 0.542 & 0.676 & 1.022 & 1.082 & {\bf 4.118} & {\bf 7.477}\\\hline
16 & 17 & 0.37 & 2.65 & 7 & 0.234 & 0.276 & 1.031 & 1.117 & 1.289 & 1.711\\
17 & 22 & 0.20 & 1.88 & 9 & 0.276 & 0.318 & 1.075 & 1.476 & 1.099 & 1.246\\
18 & 16 & 0.23 & 2.75 & 12 & 0.567 & 0.576 & 1.005 & 1.017 & 1.429 & 2.252\\
19 & 56 & 0.28 & 4.50 & 16 & 0.478 & 0.719 & 1.079 & 1.300 & 1.927 & 2.484\\
20 & 48 & 0.28 & 3.08 & 11 & 0.368 & 0.463 & 1.063 & 1.232 & 1.884 & 2.105\\\hline
21 & 12 & {\bf 0.51} & 2.93 & 6 & 0.117 & 0.150 & 1.015 & 1.072 & 1.242 & 2.197\\
22 & {\bf  60} & 0.19 & 2.09 & 11 & {\bf 0.851} & 0.908 & 1.105 & 1.655 & 1.085 & 1.194\\
23 & 18 & 0.26 & 1.68 & 6 & 0.392 & 0.464 & 1.004 & 1.030 & 1.150 & 1.453\\
24 & 30 & 0.33 & 8.16 & 25 & 0.405 & 0.373 & 1.103 & 1.368 & 2.483 & 4.903\\
25 & 10 & 0.32 & 4.99 & 16 & 0.398 & 0.391 & 1.011 & 1.054 & 1.182 & 1.924\\\hline
26 & 14 & 0.43 & 7.54 & 18 & 0.303 & 0.368 & 1.009 & 1.054 & 1.526 & 3.012\\
27 & 17 & 0.17 & 2.58 & 15 & 0.160 & 0.200 & 1.058 & 1.285 & 1.263 & 1.593\\
28 & 13 & 0.24 & 5.89 & 25 & 0.292 & 0.331 & 1.027 & 1.175 & 1.621 & 2.723\\
29 & 11 & 0.26 & 5.37 & 20 & 0.548 & 0.580 & 1.003 & 1.020 & 1.380 & 1.989\\
30 & 20 & 0.28 & 9.41 & 34 & 0.700 & {\bf 0.980} & 1.030 & 1.126 & 2.474 & 4.064\\\hline\hline
 case  &   $I$ &   $\min(\sigma_x)$ &   $\max(\sigma_x)$ &  $\displaystyle{\frac{\max(\sigma_x)}{\min(\sigma_x)}}$   & $\sigma_0(w=1)$ &  $\sigma_0(w(s))$ &  $l_{\mbox{\tiny mean}}^{\mbox{\tiny ALG$|$ML}}$   &  $l_{\mbox{\tiny max}}^{\mbox{\tiny ALG$|$ML}}$ &  $l_{\mbox{\tiny mean}}^{\mbox{\tiny 1$|$s}}$   & $l_{\mbox{\tiny max}}^{\mbox{\tiny 1$|$s}}$ \\
      & & [px] & [px] & &  [px] & [px] & & & \\ \hline
mean & 27 &       0.31 &  4.83 & 17 &  0.366 &  0.417 &  1.052 &  1.246 &  1.702 &  2.591 \\
max &     60 &    0.51 & 14.79 & 77 &  0.851 &   0.980 &  1.169 &  1.918 &  4.118 &  7.477
    \end{tabular}
    \caption{Data set B, cases 1 -- 30. Comparing the accuracy of different homography estimates. Columns: (1) Number $I$ of point pairs, (2--4) minimal and maximal standard deviations  $\min(\sigma_x)$ and $\max(\sigma_x)$ of weighted points and their ratio, (5--6) estimated variance factors for unweighted and weighted points, (7--8) mean and maximal losses when comparing the algebraic with the ML-estimate using equal weights, $l_{\mbox{\tiny mean}}^{\mbox{\tiny ALG$|$ML}}$ and  $l_{\mbox{\tiny max}}^{\mbox{\tiny ALG$|$ML}}$, and  (9--10) mean and maximal losses when comparing the unweighted and the weighted ML-estimate, $l_{\mbox{\tiny mean}}^{\mbox{\tiny 1$|$s}}$ and  $l_{\mbox{\tiny max}}^{\mbox{\tiny 1$|$s}}$\\ \\}
    \label{tab:experiment-large-1-30}
\end{table*}

\begin{figure}
    \centering
    \includegraphics[width=0.5\textwidth]{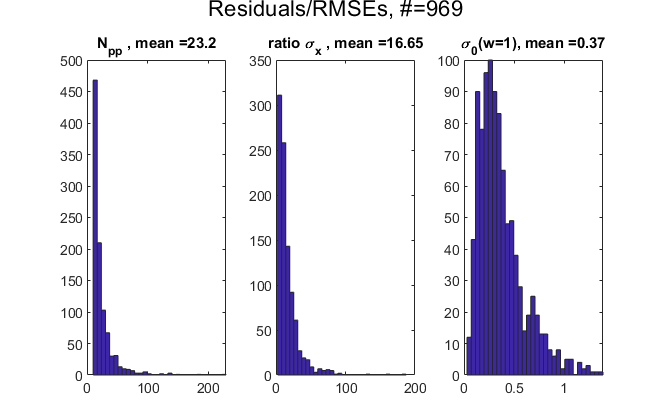}
    \caption{Data set B with 969 cases. Histograms of number of correspondences, ratio of scale dependent standard deviations of coordinates, and estimated (square rooted) variance factor $\hat {\sigma}_0$}
    \label{fig:ex-969-given}
\end{figure}
\begin{figure}
    \centering
    \includegraphics[width=0.45\textwidth]{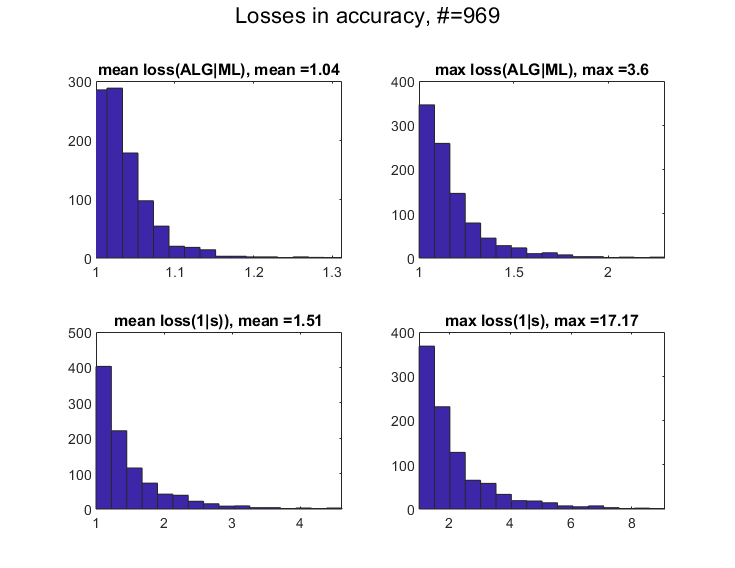}
    \caption{Data set B with 969 cases. Histograms of mean and maximum losses due to using an approximate/algebraic estimation method, and due to using equal weighting instead of scale dependent weighting }
    \label{fig:ex-969-losses}
\end{figure}

Analysing the complete data set B with 969 cases yields the results shown in the histograms of Fig \ref{fig:ex-969-given} and \ref{fig:ex-969-losses}. The maximum ratios of the scale dependent standard deviations on an average are 15.5, which appears to be quite large, however, confirming the results shown in Tab.  \ref{tab:experiment-large-1-30}. The mean variance factor is $\overline{\sigma_0^2}=(0.37 \mbox{ [px]})^2$, being consistent with earlier investigations.

Finally, we determined the {\em average residuals} for each case using the following symmetric root mean square error .
\begin{equation}
    \mbox{RMSE} = \sqrt{\frac 1 {8I} \sum_{i=1}^I |\mat x'_i-{\cal H}(\mat x_i)|^2+|\mat x_i-{\cal H}^{-1}(\mat x_i')|^2}\,,
\end{equation}
being the quadratic mean of the $\epsilon_i$ in (\ref{eq:epsilon}).

\begin{figure}
    \centering
    \includegraphics[width=0.5\textwidth]{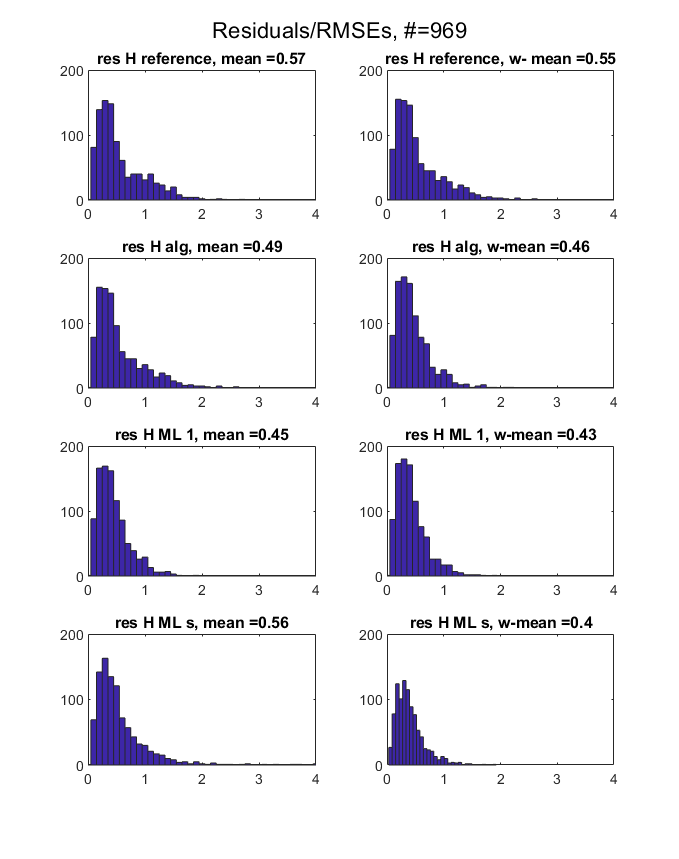}
    \caption{The residuals of the correspondences for the ({\bf columns}) reference homography, the estimated homographies using the algebraic, the unweighted ML and the scale-weighted ML estimation. {\bf Left column:} equally weighted residuals. {\bf Right column:} scale-weighted residuals. }
    \label{fig:ex-969-res}
\end{figure}

We compared the four homographies
\begin{enumerate}
    \item the reference ({\sf reference}),
    \item estimated by the algebraic minimization ({\sf alg}),
    \item estimated by the unweighted ML-estimation ({\sf ML~1}), and
    \item estimated by the scale weighted ML-estimation ({\sf ML~s}).
\end{enumerate}
The results are shown in the left column of Fig. \ref{fig:ex-969-res}.
Obviously all estimates lead to smaller residuals. Obviously, the unweighted ML-estimation leads to smaller residuals, than the weighted ML-estimate. This seems to be surprising, since one would expect the weighted ML-solution leads to better results. However, the result is consistent with theory, since the RMSE does not use any weighting, hence the unweighted ML-estimate needs to minimize the unweighted RMSE. 

If we, therefore, analyze the weighted residuals, using a weighted root mean square error 
\begin{equation}
    \mbox{RMSE}_w = \sqrt{\frac  {\sum_{i=1}^I w_i(|\mat x'_i-{\cal H}(\mat x_i)|^2+|\mat x_i-{\cal H}^{-1}(\mat x_i')|^2)} {8\sum_{i=1}^I w_i}}\,,
\end{equation}
with 
\begin{equation}
    w_i=\frac 1 {s_i^2+s_{i'}^2}\,,
\end{equation}
we obtain the histograms in the right column of Fig. \ref{fig:ex-969-res}. Now, as to be expected, the weighted residuals of the weighted ML-estimate are minimal, consistent with the theoretical expectation. Also observe, all weighted residuals are smaller than the unweighted residuals, indicating the need to weight the coordinates used for estimating the homographies.

{\small
\bibliographystyle{ieee_fullname}
\bibliography{egbib}
}

\end{document}